\documentclass{article}
\usepackage{microtype}
\usepackage{graphicx}
\usepackage{subfigure}
\usepackage{booktabs} 

\usepackage[hyperfootnotes=false]{hyperref}
\usepackage{multirow}

\usepackage{arxiv}

\usepackage{amsmath}
\usepackage{amssymb}
\usepackage{mathtools}
\usepackage{amsthm}

\usepackage{amsmath,amsfonts,bm}

\def\eqref#1{equation~\ref{#1}}

\def\1{\bm{1}}

\def\vw{{\bm{w}}}
\def\vx{{\bm{x}}}
\def\vy{{\bm{y}}}
\def\vz{{\bm{z}}}

\def\mB{{\bm{B}}}

\def\mI{{\bm{I}}}

\def\mP{{\bm{P}}}
\def\mQ{{\bm{Q}}}

\def\mX{{\bm{X}}}

\DeclareMathAlphabet{\mathsfit}{\encodingdefault}{\sfdefault}{m}{sl}
\SetMathAlphabet{\mathsfit}{bold}{\encodingdefault}{\sfdefault}{bx}{n}

\def\gD{{\mathcal{D}}}

\def\gL{{\mathcal{L}}}

\def\gN{{\mathcal{N}}}

\def\gS{{\mathcal{S}}}

\def\gtr{{\mathcal{TR}}}

\def\sR{{\mathbb{R}}}

\newcommand{\E}{\mathbb{E}}

\newcommand{\proj}{\mathrm{Proj}}

\usepackage[capitalize,noabbrev]{cleveref}
\usepackage{empheq}
\usepackage{framed}
\usepackage{enumitem}
\newlist{Aenumerate}{enumerate}{1}
\theoremstyle{plain}
\newtheorem{theorem}{Theorem}[section]
\newtheorem{proposition}[theorem]{Proposition}
\newtheorem{lemma}[theorem]{Lemma}
\newtheorem{corollary}[theorem]{Corollary}
\theoremstyle{definition}

\newtheorem{assumption}[theorem]{Assumption}
\theoremstyle{remark}

\newcommand{\abs}[1]{\left|#1\right|}
\newcommand{\defeq}{\coloneqq}

\usepackage[textsize=tiny]{todonotes}
\expandafter\def\expandafter\normalsize\expandafter{%
    \normalsize
    \setlength\abovedisplayskip{6pt}
    \setlength\belowdisplayskip{6pt}
    \setlength\abovedisplayshortskip{6pt}
    \setlength\belowdisplayshortskip{6pt}
}
\title{Theory on Forgetting and Generalization of Continual Learning}

\author{%
   Sen Lin\thanks{Equal Contribution}\\
   Department of ECE\\
   The Ohio State University\\
   \texttt{lin.4282@osu.edu}
   \And
   Peizhong Ju$^*$\\
  Department of ECE\\
   The Ohio State University\\
   \texttt{ju.171@osu.edu}
   \And
   Yingbin Liang\\
   Department of ECE\\
   The Ohio State University\\
   \texttt{liang.889@osu.edu}
   \And
   Ness Shroff\\
   Department of ECE\\
   The Ohio State University\\
   \texttt{shroff.11@osu.edu}
}

\begin{document}
\maketitle

\begin{abstract}
Continual learning (CL), which aims to learn a sequence of tasks, has attracted significant recent attention. However, most work has focused on the experimental performance of CL, and theoretical studies of CL are still limited. In particular, there is a lack of understanding on what factors are important and how they affect  ``catastrophic forgetting" and  generalization performance.
To fill this gap, our theoretical analysis, under overparameterized linear models, provides the first-known explicit form of the expected forgetting and generalization error.
Further analysis of such a key result yields a number of theoretical explanations about how overparameterization, task similarity, and task ordering affect both forgetting and generalization error of CL.
More interestingly, by conducting experiments on real datasets using deep neural networks (DNNs), we show that some of these insights even go beyond the linear models and can be carried over to practical setups.
In particular, we use concrete examples to show that our results not only explain some interesting empirical observations in recent studies, but also motivate better practical algorithm designs of CL.
\end{abstract}

\section{Introduction}
\label{introduction}

Continual learning (CL) \cite{parisi2019continual} is a learning paradigm where an agent needs to continuously learn a sequence of tasks. To resemble the extraordinary lifelong learning capability of human beings, the agent is expected to learn new tasks more easily based on  accumulated knowledge from old tasks, and further improve the learning performance of old tasks by leveraging the knowledge of new tasks. The former is referred to as forward knowledge transfer and the latter as backward knowledge transfer. One major challenge herein is the so-called {\em catastrophic forgetting} \cite{mccloskey1989catastrophic}, i.e., the agent easily forgets the knowledge of old tasks when learning new tasks.

Although there have been significant efforts in experimental studies (e.g., \cite{kirkpatrick2017overcoming, chaudhry2018efficient, yoon2020scalable, doan2021theoretical, evron2022catastrophic}) to address the forgetting issue, the theoretical understanding of CL is still in the early stage, where only a few attempts have emerged  recently, e.g., \cite{yin2020optimization,bennani2020generalisation,doan2021theoretical,evron2022catastrophic} (see a more detailed discussion about the previous theoretical studies of CL in Section~\ref{sec:relatedwork}).
However, none of these existing theoretical results provide an explicit characterization of forgetting and generalization error, that only depends on fundamental system parameters/setups (e.g., number of tasks/samples/parameters, noise level, task similarity/order). Thus, our work here provides the first-known  explicit theoretical result, which enables us to comprehensively understand which factors are relevant and how they (precisely) affect forgetting and generalization error of CL.  



Our main contributions can be summarized as follows.

First, we provide theoretical results on the expected value of forgetting and overall generalization error in CL, under a linear regression setup with \emph{i.i.d.} Gaussian features and noise. The expression of our results is in an explicit form that captures a clear dependency on various system parameters/setups. 
Note that analyzing overparameterized linear models are important in their own right and also, as demonstrated in many recent works, are a first step towards understanding the generalization performance of DNNs, e.g.,
\cite{belkin2018understand,bartlett2020benign,ju2020overfitting,muthukumar2020harmless,hastie2022surprises}.

Second, we investigate the impact of overparameterization, 
task similarity, and task ordering
on both forgetting and 
generalization error of CL, which reveals the following important insights: 1)  Both forgetting and generalization error can benefit from more parameters in the overparameterized regime. Moreover, benign overfitting exists and is easier to observe with large noise and/or low task similarity. 
2) In terms of the impact of task similarity, we show that the generalization error always decreases when tasks become more similar, whereas this `monotonicity' does not always hold for  forgetting. Surprisingly, forgetting can even decrease when tasks are less similar under certain scenarios. 3) In order to minimize forgetting, the optimal task order should diversify the learning tasks in the early stage and learn more dissimilar tasks adjacently. This is also corroborated by some special scenarios where the tasks can be divided into multiple categories, and the optimal task order therein alternatively learns tasks from different categories.

Last but not least, we  show that our findings for the linear models are applicable to and can also guide the algorithm designs for CL in practice, by conducting experiments on real datasets with DNNs. Specifically, 
our analysis of the impact of task similarity is clearly corroborated by the experimental results, which further sheds light on the recent observations \cite{ramasesh2020anatomy,lee2021continual,evron2022catastrophic} that `intermediate task similarity' leads to the worst forgetting in the two-task setup.  
Experimental results on the impact of task ordering are also consistent with our findings in  linear models. More interestingly, inspired by our analysis of knowledge transfer in linear models, we slightly modify a previous method \cite{lin2022trgp} on leveraging task correlation to facilitate forward knowledge transfer, and show that better performance can be achieved by counting more on fresher old tasks. These encouraging results  corroborate the benefits of studying the overparameterized linear models to fundamentally demystify CL.

\section{Related Work}\label{sec:relatedwork}

\textbf{Empirical studies in CL.} CL has attracted much attention in the past decade, and a vast amount of empirical methods have been proposed to address catastrophic forgetting. In general, the existing methods can be divided into three categories: (1) \emph{Regularization-based methods} (e.g., \cite{kirkpatrick2017overcoming, aljundi2018memory, liu2022continual}), which regularize the modifications on the important weights to old tasks when learning the new task; (2) \emph{Parameter-isolation based methods} (e.g., \cite{serra2018overcoming, yoon2020scalable, yang2021grown}), which learn a mask to fix the important weights to old tasks during the new task learning and further expand the neural network when needed; (3) \emph{Memory-based methods}, which either store and replay data of old tasks when learning the new task, i.e., experience-replay based methods (e.g., \cite{chaudhry2018efficient, riemer2018learning, jin2021gradient}), or store the gradient information of old tasks and learn the new task in the orthogonal direction to old tasks without data replay, i.e., orthogonal-projection based methods (e.g., \cite{farajtabar2020orthogonal, saha2021gradient, lin2022trgp}).

\textbf{Theoretical studies in CL.} 
Specifically, \cite{bennani2020generalisation} and \cite{doan2021theoretical} analyzed generalization error and forgetting for the orthogonal gradient descent (OGD) approach \cite{farajtabar2020orthogonal} based on NTK models, and further proposed variants of OGD to address forgetting. \cite{yin2020optimization} proposed a unified framework for the performance analysis of regularization-based CL methods, by formulating them as a second-order Taylor approximation of the loss function for each task. \cite{asanuma2021statistical} and \cite{lee2021continual} studied CL in the teacher-student setup to characterize the impact of task similarity on  forgetting performance. \cite{cao2022provable} and \cite{li2022provable} investigated continual representation learning with dynamically expanding feature spaces, and developed provably efficient CL methods with a characterization of the sample complexity. \cite{chen2022memory} characterized the lower bound of memory in CL using the PAC framework. By investigating the information flow between neural network layers,  \cite{andle2022theoretical} analyzed the selection of frozen filters based on layer sensitivity to maximize the performance of CL. 
Nevertheless, none of these existing works  show an explicit form of forgetting and generalization error, that only depends on fundamental system parameters/setups (e.g., number of tasks/samples/parameters, noise level, task similarity/order). In contrast, our work is the first one to provide such an explicit theoretical result, which enables us to comprehensively understand what factors (and how they) affect the forgetting and generalization performance of CL.

The most relevant study to our work is \cite{evron2022catastrophic}, which also studied CL in overparameterized linear models. However, our work is quite different from \cite{evron2022catastrophic}: (1) We study and provide the exact forms of both forgetting and generalization error based on the testing loss, while \cite{evron2022catastrophic} only evaluated forgetting using the training data; (2) Our results characterize the performance of CL in a comprehensive way, through investigating 
how overparameterization, task similarity and task ordering  affect both forgetting and generalization error, while \cite{evron2022catastrophic} only studied the upper bound of catastrophic forgetting under specific task orderings; (3) Unlike \cite{evron2022catastrophic}, our study is able to explain recent phenomena and guide the algorithmic development in CL with DNN.


\textbf{Studies about generalization performance on overparameterized models (benign overfitting).} 
DNNs are usually so overparameterized that can completely fit all training samples, yet they can still generalize well on unseen test data. This seems to contradict the classical knowledge of \emph{bias-variance trade-off}. 
As a first step of understanding this mystery, the ``benign overfitting'' or ``double-descent'' phenomenon\footnote{i.e., test error decreases again in the overparameterized region with more parameters, so the overfitting is benign for the generalization performance.} has been discovered and studied for overfitted solutions of single-task linear regression. For example, some work discovered and studied double-descent with min $\ell_2$-norm overfitted solutions \cite{belkin2018understand, belkin2019two,bartlett2019benign,	hastie2019surprises,muthukumar2019harmless} or min $\ell_1$-norm overfitted solutions \cite{mitra2019understanding,ju2020overfitting}, while using simple features such as Gaussian or Fourier features.
Some other recent work studied the overfitted generalization performance by adopting features that approximate shallow neural networks, for example, random feature (RF) models \cite{mei2019generalization}, two-layer neural tangent kernel (NTK) models \cite{arora2019fine,satpathi2021dynamics,ju2021generalization}, and three-layer NTK models \cite{ju2022generalization}. All of these studies considered only  a single task. In contrast, our work focuses on CL with a sequence of tasks, which brings in many new variables such as task similarity and task ordering.

\section{Continual Learning in Linear Models}

Consider the standard CL setup where a sequence of tasks $\mathbb{T}=\{1,...,T\}$ arrives sequentially in time.
 
\textbf{Ground truth.} We consider a linear ground truth \cite{belkin2018understand,evron2022catastrophic} for each task. Specifically, for task $t$, the output $y\in \mathbb{R}$ is given by
\begin{align}\label{eq.ground_truth_origin}
    y_t=\hat{\vx}_t^{\top}\hat{\vw}_{t}^* + z_t,
\end{align}
where $\hat{\vx}_t\in\mathbb{R}^{s_t}$ denotes the feature vector,  $\hat{\vw}_{t}^*\in \mathbb{R}^{s_t}$ denotes the model parameters, and $z_t$ is the random noise.  Here $s_t$ denotes the number of features of ground truth (i.e., the number of true features). 
In practice, true features are unknown in advance. Therefore, when choosing a model to learn a certain task, people usually choose more features than enough such that all possible features are included. We write this formally into the following assumption\footnote{When \cref{as.enough_features} does not hold, the derivation techniques for \cref{thm:exactform} in the next section still hold with a minor modification that treats the missing features as noise.}.
\begin{assumption}\label{as.enough_features}
We index all possible features by $1,2,\cdots$.
Let $\mathcal{W}$ denote the set of indices of all the chosen features in the model to be trained, with cardinality $\abs{\mathcal{W}}=p$. Let $\mathcal{S}_t$ denote the set of indices of $t$-th task's true features, with cardinality $\abs{\mathcal{S}_t}=s_t $. We assume that $\bigcup_{t\in \mathbb{T}}\mathcal{S}_t \subseteq \mathcal{W}$.  
\end{assumption}

We next define an expanded ground-truth vector $\vw_t^*\in \mathbb{R}^p$ that expands the original ground-truth vector $\hat{\vw}_t^*$ from dimension $s_t$ to dimension $p$ by filling zeros in the positions $\mathcal{W}\setminus \mathcal{S}_t$. Let $\vx_t$ be the corresponding features for $\vw_t^*$. Therefore, the ground truth \cref{eq.ground_truth_origin} can be rewritten as
\begin{align}\label{eq.true_model}
    y_t=\vx_t^{\top}\vw_t^* + z_t.
\end{align}



\textbf{Data.} For each task $t\in \mathbb{T}$, the training dataset is denoted as $\mathcal{D}_t=\{(\vx_{t,j}, y_{t,j})\in \mathbb{R}^{p}\times \mathbb{R}\}_{j\in [n_t]}$ with sample size $n_t$. By stacking the training data as $\mX_t\defeq [\vx_{t,1}\ \vx_{t,2}\ \cdots\ \vx_{t,n_{t}}]\in \mathbb{R}^{p\times n_{t}}$ and $\vy_t\defeq [y_{t,1}\ y_{t,j}\ \cdots\ y_{t,n_{t}}]^{\top}\in \mathbb{R}^{n_{t}\times 1}$, \cref{eq.true_model} can be written as
\[\vy_t = \mX_t^{\top}\vw_t^* + \vz_t.\]

To simplify our analysis,  we consider \emph{i.i.d.} Gaussian features and noise, which is stated in the following assumption.
\begin{assumption}\label{as.Gaussian}
Each element of $\mX_t$ for all $t\in \mathbb{T}$ follows standard Gaussian distribution $\gN(0,1)$ and is independent of each other. The noise $\vz_t\sim \gN(\bm{0}, \sigma_t^2 \mI_p)$ and is independent of each other for all $t\in \mathbb{T}$, where $\sigma_t\geq 0$ denotes the noise level.
\end{assumption}

\textbf{Learning procedure.}
We train the model parameters $\vw$ for each task sequentially.
Let $\vw_t$ denote the result after training for task $t$, which is also the initial point in the model training for task $t+1$. Let $\vw_0=\bm{0}$, i.e., task~1 starts training from zero. 
For each task $t$, the training loss is defined by mean-squared-error (MSE) with respect to (w.r.t.) $(\mX_t,\vy_t)$:
\begin{align}\label{eq:loss}
   \gL^{tr}_t(\vw, \gD_t)=\frac{1}{n_t}\|(\mX_t)^{\top}\vw-\vy_t \|_2^2.
\end{align}
When underparameterized (i.e., $n_t\leq p$), minimizing \cref{eq:loss} has a unique solution (with probability 1).
When overparameterized (i.e., $p>n_t$), minimizing \cref{eq:loss} has an infinite number of solutions that make \cref{eq:loss} zero. Among all overfitted solutions, we are particularly interested in the one corresponding to the convergent point of stochastic gradient descent (SGD) for minimizing \cref{eq:loss}. In fact, it can be shown that such an overfitted solution has the smallest $\ell_2$-norm of the change of parameters \cite{gunasekar2018characterizing}. In other words, $\vw_t$ corresponds to the solution to the following optimization problem:
\begin{align}\label{eq:op}
    \min_{\vw}~~ \|\vw-\vw_{t-1}\|_2,~~~
    s.t.~~ (\mX_t)^{\top}\vw=\vy_t.
\end{align}
The constraint in \cref{eq:op} implies that the training loss is exactly zero (i.e., overfitted).

\textbf{Performance evaluation.} For the described linear system, we  use $\gL_t(\vw)$ to denote the model error\footnote{It can be proved that the model error we defined here is equivalent to the mean-squared-error on noise-free test data.} for task $t$:
\begin{align}\label{eq:def_model_error}
    \gL_t(\vw)=\|\vw-\vw_t^*\|^2,
\end{align}
which characterizes the generalization performance of $\vw$ on task~$t$.
As is standard in the empirical studies of CL, e.g., \cite{chaudhry2018efficient, lin2022trgp}, we evaluate the performance of CL on two key metrics, forgetting and overall generalization error, defined as below:

(1) \emph{Forgetting:} It measures how much `knowledge' of old tasks has been forgotten after learning the current task. Specifically, after learning task $t\in [2,T]$, the average  forgetting over all old tasks $i\in[1,t-1]$ is defined as:
    \begin{align}\label{eq:forget_linear}
       F_t=\frac{1}{t-1}\sum_{i=1}^{t-1} (\gL_i(\vw_t)-\gL_i(\vw_i)).
    \end{align}
    
    In \cref{eq:forget_linear}, $\gL_i(\vw_t)-\gL_i(\vw_i)$ denotes the performance difference between $\vw_i$ (the result after training task $i$) and $\vw_t$ (the result after training task $t$) on test data of task $i$.
    
(2) \emph{Overall generalization error:} 
    We evaluate the model generalization performance of the final task model $\vw_T$ in terms of the average model error over all tasks:   
\begin{align}\label{eq:generalization}
        G_T=\frac{1}{T}\sum_{i=1}^T \gL_i(\vw_T).
    \end{align}
    
It is worth noting that the forgetting defined in \cite{evron2022catastrophic} is based on the training loss, which consequently ignores the generalization performance of the learned models for old tasks. Such a definition is not only inconsistent with the evaluation metric in empirical studies, but also insufficient to capture the backward knowledge transfer because the value of forgetting therein can not be negative.

We further simplify the current setup by assuming that each task has the same number of training samples as well as the same noise level $\sigma$, stated as follows.
\begin{assumption}\label{as:same_train_num_and_noise}
$n_t=n$ and $\sigma_t=\sigma$ for all $t\in\mathbb{T}$.
\end{assumption}
Note that \cref{as:same_train_num_and_noise} is adopted only to make our results (which will be shown in the next section) easy to interpret. In fact, our analysis can be easily generalized to the situation when \cref{as:same_train_num_and_noise} does not hold.
\section{Main Results and Interpretations}



\begin{figure*}[ht]
\centering

\includegraphics[width=0.95\textwidth]{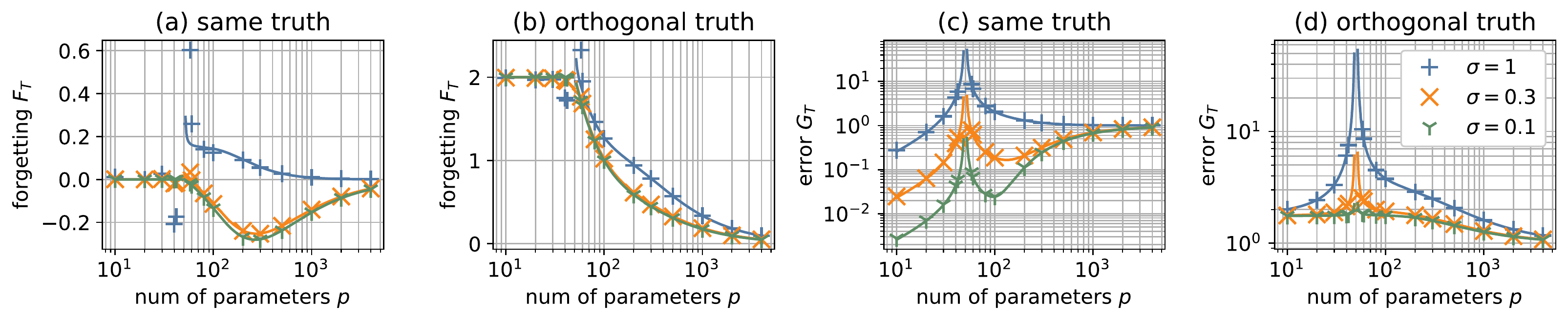}

\caption{The trend of forgetting and overall generalization error w.r.t. the number of model parameters, where $T=8$, $n=50$, $\hat{\vw}_t^*\in \mathbb{R}^{10}$ and $\|\hat{\vw}_t^*\|^2=1$ for all $t\in \mathbb{T}$.  The ground truths are the same for all tasks in
Subfigures~(a) and (c), but are orthogonal in Subfigures~(b) and (d)  where $\hat{\vw_t^*}$ equals to $t$-th standard basis for all $t\in \mathbb{T}$. The discrete points indicated by markers are calculated by simulation and are the average of $300$ random simulation runs. The curves are drawn by the theoretical expressions in \cref{thm:exactform} and \cref{thm:underparameterized}.}\label{fig:linear}

\end{figure*}

Although we use linear models, in order to provide hints on understanding DNNs that are usually heavily overparameterized, we are particularly interested in the performance of CL in the overparameterized region ($p>n$), where we define the overparameterized ratio as $r\defeq 1-\frac{n}{p}$. For ease of exposition,
we define the following coefficients that will appear in our main theorem:
\begin{align}\label{eq:cij}
    c_{i,j}\defeq(1-r)\left(r^{T-i}-r^{j-i}+r^{T-j}\right),
\end{align}
where $1\leq i<j\leq T$ are the indices of tasks.
Now we are ready to state our main theorem that characterizes the expected value of forgetting and overall generalization error:
\begin{theorem}\label{thm:exactform}
    When $p\geq n+2$, we must have
    \begin{align}\label{eq:forget_close}
        \mathbb{E}[F_T]=\frac{1}{T-1}\sum_{i=1}^{T-1}\Big[\underbrace{(r^T-r^i)\|\vw_i^*\|^2}_{\text{Term~F1}}+\underbrace{\sum_{j>i}^T c_{i,j}\|\vw_i^*-\vw_j^*\|^2}_{\text{Term~F2}}
        +\underbrace{\frac{p\sigma^2}{p-n-1}(r^i-r^T)}_{\text{Term~F3}}
        \Big]
    \end{align}
    \begin{align}\label{eq:general_close}
        \mathbb{E}[G_T]=\underbrace{\frac{r^T}{T}\sum_{i=1}^T \|\vw_i^*\|^2}_{\text{Term~G1}}+\underbrace{\frac{1}{T}\sum_{i=1}^T \frac{nr^{T-i}}{p}\sum_{k=1}^T\|\vw_k^*-\vw_i^*\|^2}_{\text{Term~G2}}
        +\underbrace{\frac{p\sigma^2}{p-n-1}\left(1-r^T\right)}_{\text{Term~G3}}.
    \end{align}
\end{theorem}
To the best of our knowledge, \cref{thm:exactform} is the first result that establishes the closed forms of forgetting and overall generalization error of CL in overparameterized linear models. In the rest of the paper, we will see that \cref{thm:exactform} not only describes how CL performs on the linear system but also provides guidance on applying CL in practice that DNNs and real-world datasets. The proof of \cref{thm:exactform} is in \cref{proof:thm:exactform}. We also verify the correctness of \cref{thm:exactform} in \cref{fig:linear} where discrete points indicated by markers in \cref{fig:linear} (drawn by simulations) are very close to the curves (drawn by \cref{thm:exactform} and \cref{thm:underparameterized}).

We can further simply \cref{eq:forget_close} and \cref{eq:general_close} by only considering two tasks, so as to better understand \cref{thm:exactform}. The result is shown in the following corollary, which clearly characterizes the dependence on task similarity and different system parameters.

\begin{corollary}
When $T=2$ and $p\geq n+2$, we must have
\begin{align}
   \mathbb{E}[F_2]=&(r^2-r)\|\vw_1^*\|^2+\frac{n}{p}\|\vw_2^*-\vw_1^*\|^2+\frac{nr\sigma^2}{p-n-1},\label{eq:forget_two}\\
    \mathbb{E}[G_2]=&\frac{r^2}{2}\left(\|\vw_1^*\|^2+\|\vw_2^*\|^2\right) + \frac{1-r^2}{2}\|\vw_1^*-\vw_2^*\|^2
    +\frac{p\sigma^2(1-r^2)}{p-n-1}.\label{eq:generalization_two}
\end{align}
\end{corollary}

Based on  \cref{thm:exactform}, we will provide insights on the following three aspects.

(1) \emph{Overparameterization (\cref{subsec:overparameterization}).} In order to understand the generalization power of overfitted machine learning models, much attention has focused (e.g., \cite{belkin2018understand, ju2020overfitting, hastie2022surprises}) on studying the impact of overparameterization on single-task learning, whereas how overparameterization affects the performance of CL still remains unclear. Fortunately, the exact forms  in \cref{thm:exactform} provide a way to directly evaluate the impact of overparameterization and the random noise on both forgetting and generalization error in CL.

(2) \emph{Task similarity (\cref{subsec:correlation}).} Both forgetting and generalization error depend on the optimal model gap between any two tasks , i.e., $\|\vw_k^*-\vw_i^*\|^2$ for any task $k$ and $i$, which defines the task similarity in this work (smaller gap means higher similarity).  Understanding the impact of task similarity is helpful to not only explain empirical observations but also guide better designs of CL in practice.

(3) \emph{Task ordering (\cref{subsec:order}).} Given a fixed set of tasks in CL, the learning order of the task sequence clearly plays an important role in affecting both $\E[F_T]$ and $\E[G_T]$, through the task order-dependent coefficients, e.g., $c_{ij}$ in \cref{eq:forget_close} and $r^{T-i}$ in \cref{eq:general_close}. For example, suppose $\|\vw_i^*\|^2$ is the same for all $i\in \mathbb{T}$, the optimal task ordering to minimize the generalization error  is to learn the tasks in a decreasing order of $\sum_{k=1}^T\|\vw_k^*-\vw_i^*\|^2$, i.e., $i<j$ if $\sum_{k=1}^T\|\vw_k^*-\vw_i^*\|^2>\sum_{k=1}^T\|\vw_k^*-\vw_j^*\|^2$. Intuitively, the most dissimilar task should be learnt first in this case. Investigating the impact of task ordering is particularly valuable when the agent can control the task order in CL, in the same spirit of curriculum learning \cite{bengio2009curriculum}.

In what follows, we will delve into the impact of those three crucial factors  in order to provide a comprehensive understanding of CL in the linear models.

\subsection{The impact of overparameterization}\label{subsec:overparameterization}

In this subsection, we show some insights about the impact of overparameterization. Specially, we will discuss what happens when $p$ changes under a fixed $n$.

\textbf{1) More parameters can lead to zero forgetting and alleviate the negative impact of task dissimilarity on generalization error.} As shown in \cref{thm:exactform},  when $p\to \infty$, we can have that $\mathbb{E}[F_T]\to 0$ and Term G2 also approaches zero.  In some special cases, we can further show that Term G2 is monotonically decreasing w.r.t. $p$.
A more detailed discussion can be found in \cref{app:over}.



\textbf{2) Benign overfitting exists and is easier to observe with large noise and/or low task similarity.}
As we introduced in related work, benign overfitting has recently been discovered and studied in linear models as a first step towards understanding why DNNs can still generalize well even when heavily overparameterized. The concept of ``benign overfitting'' and ``double-descent'' is initially proposed for only a single task. We now show that such a phenomenon also exists in CL where there exists a sequence of tasks.

Notice that \cref{thm:exactform} is for the overparameterized region.
For a precise comparison between the performance of overfitting and underfitting, we present the theoretical result of the underparameterized region in the following theorem.
\begin{theorem}\label{thm:underparameterized}
When $n\geq p+2$, we must have
\begin{align*}
    \mathbb{E}[F_T]=&\frac{1}{T-1}\sum_{i=1}^{T-1}\|\vw_T^*-\vw_i^*\|^2,\\
    \mathbb{E}[G_T]=&\left(\frac{1}{T}\sum_{i=1}^{T-1}\|\vw_T^*-\vw_i^*\|^2\right) + \frac{ p\sigma^2}{n - p - 1}.
\end{align*}
\end{theorem}
We provide an intuitive explanation and rigorous proof of \cref{thm:underparameterized} in  \cref{proof:underparameterized}. 
As shown in \cref{thm:underparameterized}, $\E[G_T]$ becomes larger when the noise level $\sigma$ is larger, and both $\E[F_T]$ and $\E[G_T]$ become larger when tasks are less similar (i.e., when $\sum_{i=1}^{T-1}\|\vw_T^*-\vw_i^*\|^2$ is larger). In contrast, in the overfitted situation, Term~F2 and Term~G2 in \cref{thm:exactform} (corresponding to task similarity), Term~F3 and Term~G3 (corresponding to noise) will go to zero when $p\to \infty$. This indicates that when the noise level is high and/or task similarity is low, the performance of CL in  the overparameterized situation is more likely to be better than that in the underparameterized situation, i.e., benign overfitting exists and is easier to observe.
This can be observed from \cref{fig:linear}. For example, the blue curve with markers ``$+$'' corresponds to the largest noise (compared with other curves in \cref{fig:linear}(d)) and the lowest task similarity (compared with \cref{fig:linear}(c)), and it has the deepest descent curve in the overparameterized region ($p>50=n$). This observation indicates that benign overfitting is easier to observe with larger noise and lower task similarity.

\textbf{3) A descent floor sometimes exists on forgetting and generalization error, especially when tasks are similar and noise is low.} In \cref{eq:forget_two}, the term $(r^2-r)\|\vw_1^*\|^2$ first decreases and then increases as $p$ increases from $n$ to $\infty$ (i.e., $r$ increases from $0$ to $1$), while the remaining two terms decrease as $p$ increases. Thus, when $\|\vw_2^*-\vw_1^*\|^2$ (task similarity) and $\sigma^2$ (noise level) are relatively small, the trend of $F_2$ w.r.t. $p$ will be dominated by the first term, where a descent floor of forgetting exists. In the right-hand-side of \cref{eq:generalization_two}, the first term increases as $p$ increases, while the rest two terms decrease as $p$ increases. Taking the derivative of \cref{eq:generalization_two} on $p$, we have
\begin{align*}
    \frac{\partial \mathbb{E}[G_2]}{\partial p}=\frac{2nr{\vw_1^*}^{\top}\vw_2^*}{p^2}
    - \sigma^2\left(\frac{(n+1)(1-r^2)}{(p-n-1)^2}+\frac{2nr}{(p-n-1)p}\right).
\end{align*}
Here, since $\frac{1}{p-n-1}$ is very large when $p$ is close to $n$, while decreasing to zero when $p\to\infty$, we can tell that when $\sigma^2$ is relatively small w.r.t. ${\vw_1^*}^{\top}\vw_2^*$, $\frac{\partial \mathbb{E}[G_2]}{\partial p}$ will be positive and then negative as $p$ increases from $n+2$ to $\infty$. In other words, if these two tasks have a positive correlation (i.e., ${\vw_1^*}^{\top}\vw_2^*>0$) and noise is small, there exists a descent floor w.r.t. $p$ on $\E[G_2]$. Such a phenomenon can exist in other setups besides the special case of $T=2$. For example, in \cref{fig:linear}(a)(c) where the ground truth for each task is exactly the same, we can observe a descent floor for the small noise cases $\sigma=0.3$ and $0.1$ (i.e., orange and green curves with markers ``$\times$'' and ``Y'', respectively).

\subsection{The impact of task similarity}
\label{subsec:correlation}

\textbf{Generalization error monotonically decreases with task similarity whereas forgetting may not.}
Based on \cref{thm:exactform}, it can be seen that the generalization error $G_T(\vw_T)$ decreases when $\|\vw^*_k-\vw^*_i\|^2$ for any two different tasks $k$ and $i$ decreases, because of the positive coefficients in Term G2 in \cref{eq:general_close}. Intuitively, the generalization error of CL will be smaller if the tasks are more similar with each other. In  contrast, the forgetting $F_T$ may not  change monotonically with $\|\vw^*_k-\vw^*_i\|^2$, because the coefficients $c_{ij}$ in Term F2 in \cref{eq:forget_close} can be negative. To verify this result, we consider two different scenarios.

(1) Consider the case where $T=2$. In \cref{eq:forget_two}, $\|\vw_2^*-\vw_1^*\|^2$ captures the task similarity between tasks 1 and 2 in terms of the optimal task models. It is clear that forgetting increases with $\|\vw_2^*-\vw_1^*\|^2$, i.e., less forgetting when the two tasks are more similar.

(2) Consider the case where $T=4$. We first assume that $\|\vw_i^*\|^2  = w$ for any task $i\in [1, 4]$ considering the overparameterized models \cite{evron2022catastrophic}. Suppose that task 1 and task 2 share the same set of true features,  which is orthogonal to the feature set of both task 3 and task 4,  i.e., $\gS_1=\gS_2$ and $\gS_1\cap (\gS_3\cup\gS_4)=\emptyset$. Note that
\begin{align*}
    \|\vw_i^*-\vw_j^*\|^2=\|\vw_{i}^*\|^2+\|\vw_{j}^*\|^2-2\langle \vw_{i}^*, \vw_{j}^*\rangle
\end{align*}
where $\langle \vw_{i}^*, \vw_{j}^*\rangle=0$ if $\gS_i\cap \gS_j=\emptyset$. Therefore, we can control the value of $\|\vw_1^*-\vw_2^*\|^2$ by changing $\vw_2^*$, without affecting the value of $\|\vw_i^*-\vw_j^*\|^2$ for any pair of $\{i,j\}\neq\{1, 2\}$. Based on \cref{thm:exactform}, it can be shown that $c_{1,2}<0$, such that increasing $\|\vw_1^*-\vw_2^*\|^2$, i.e., the tasks become less similar, will surprisingly decrease forgetting.





\subsection{The impact of task ordering}
\label{subsec:order}

In order to investigate the impact of task ordering on the performance of CL, we assume that $\|\vw_t^*\|^2=w$ for every task $t\in \mathbb{T}$. By ignoring the task order-independent terms in \cref{eq:forget_close} and \cref{eq:general_close}, we focus on the task order-dependent terms, i.e., Term F2 and Term G2.

\textbf{1) Optimal task ordering of minimizing  forgetting tends to arrange  dissimilar tasks adjacently in the early stage of the sequence.}
As shown in Term F2, the optimal task order to minimize forgetting closely hinges upon the value of $c_{i,j}$. Based on \cref{eq:cij}, $c_{i,j}$ is  smaller when (1) $i$ and $j$ are smaller and (2) they are closer. Intuitively, this implies that tasks with larger $\|\vw_i^*-\vw_j^*\|^2$ should be learnt adjacently with higher priority in CL, in order to minimize the impact of the task dissimilarity on the value of $\Tilde{F}_T(\vw_T)$.  However, finding the optimal task order for the general case is highly nontrivial due to the complex coupling across $\|\vw_i^*-\vw_j^*\|^2$ for different tasks. To verify the implication above and better understand the structure of the optimal task order, we study several special cases of the task setups.

(1) \emph{[Special case I: One vs Many]~} There are two different categories of tasks, where tasks in the same category have the same optimal model; among the entire task set, one special task belongs to Category I while the other tasks belong to Category II.
In this case, the optimal task order is captured by the optimal learning order of the special task in Category I.
We have the following result to characterize the optimal task order for Special case I.
\begin{proposition}\label{pro:order1}
    Let $i^*\in [1, T]$ denote the optimal order of the special task in Category I to minimize forgetting. Suppose $p\geq n+2$. Then 1) $i^*$ can take any integer value between 2 and $\frac{T}{2}$, depending on the value of $\frac{n}{p}$; 2) $i^*$ is non-decreasing with $\frac{n}{p}$.
\end{proposition}
As indicated by \cref{pro:order1}, the special task will be learnt in the first half of the sequence, such that the task diversity in the first half is always larger than in the second half. Besides, with the model capacity increasing ($\frac{n}{p}\rightarrow 0$), the order of the special task will move towards the beginning of the sequence, because 1) the model is less concerned about the special task since it is powerful enough to learn different features and 2) the model focuses on the performance of the majority and
seeks to learn more tasks from Category II at the end of the sequence for better performance.

\begin{figure*}[ht]

    \centering
    \subfigure[]{\includegraphics[width=0.21\textwidth]{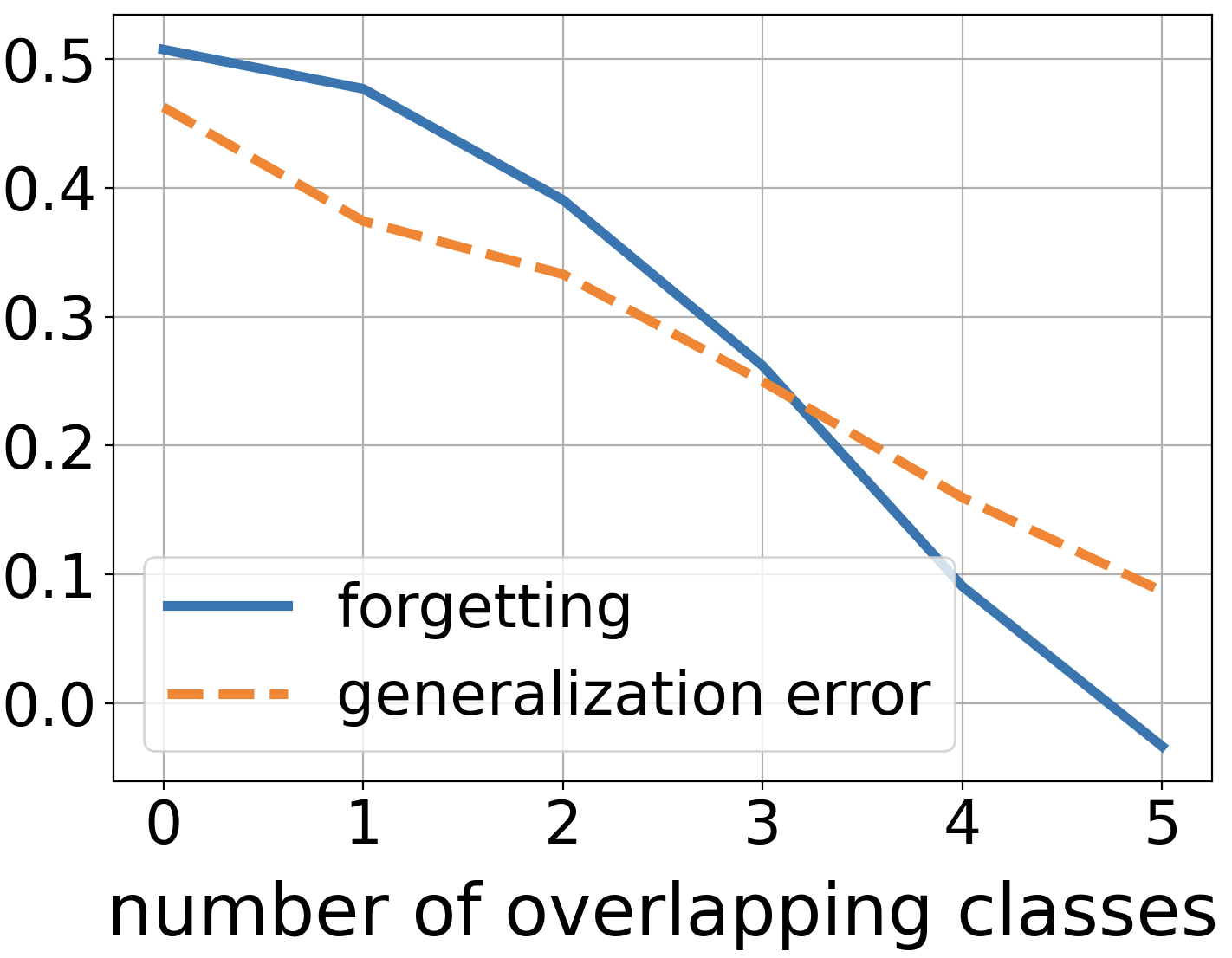}
    \label{fig:d2}
    }
    \subfigure[]{\includegraphics[width=0.21\textwidth]{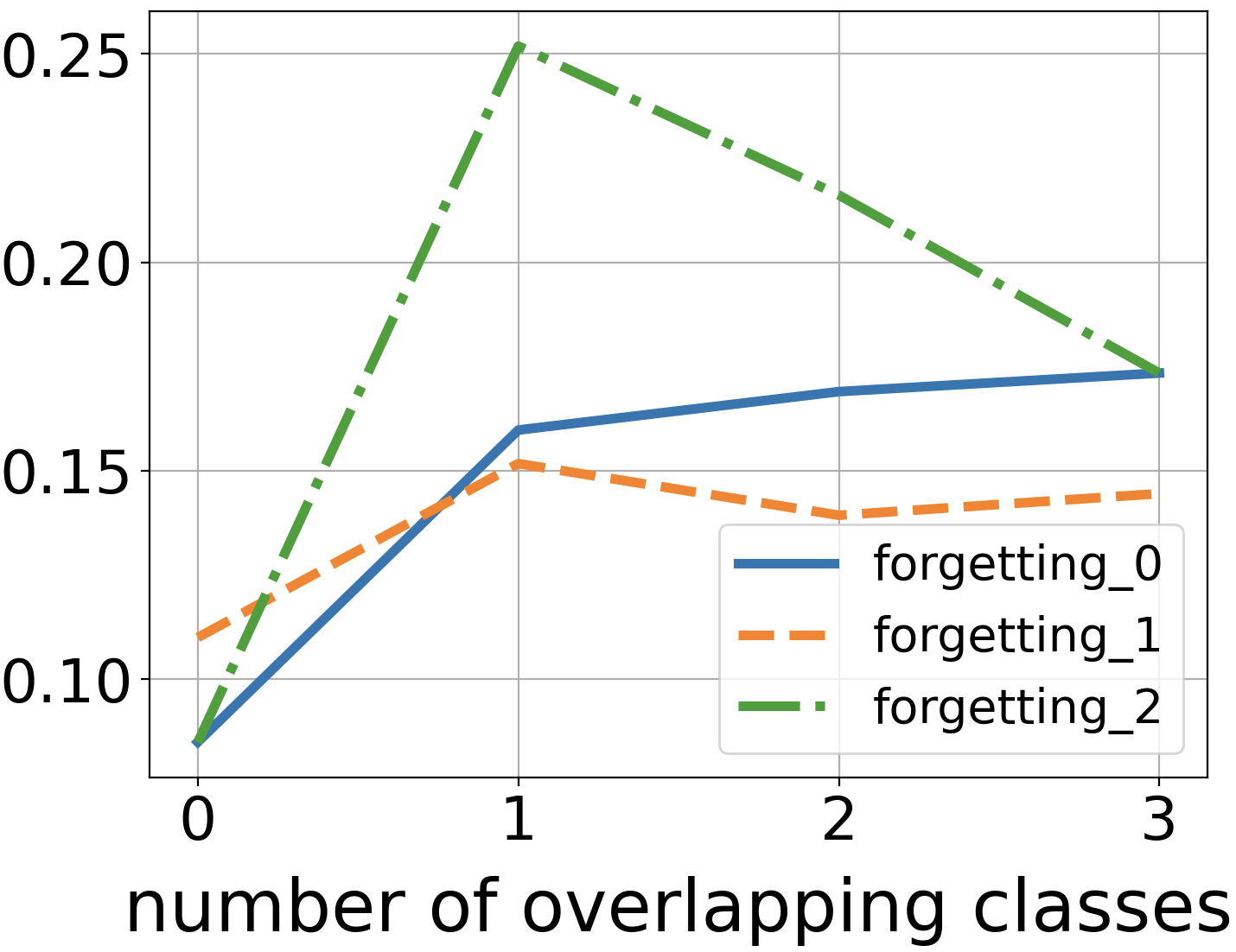}
    \label{fig:d4}
    }
    \subfigure[]{\includegraphics[width=0.21\textwidth]{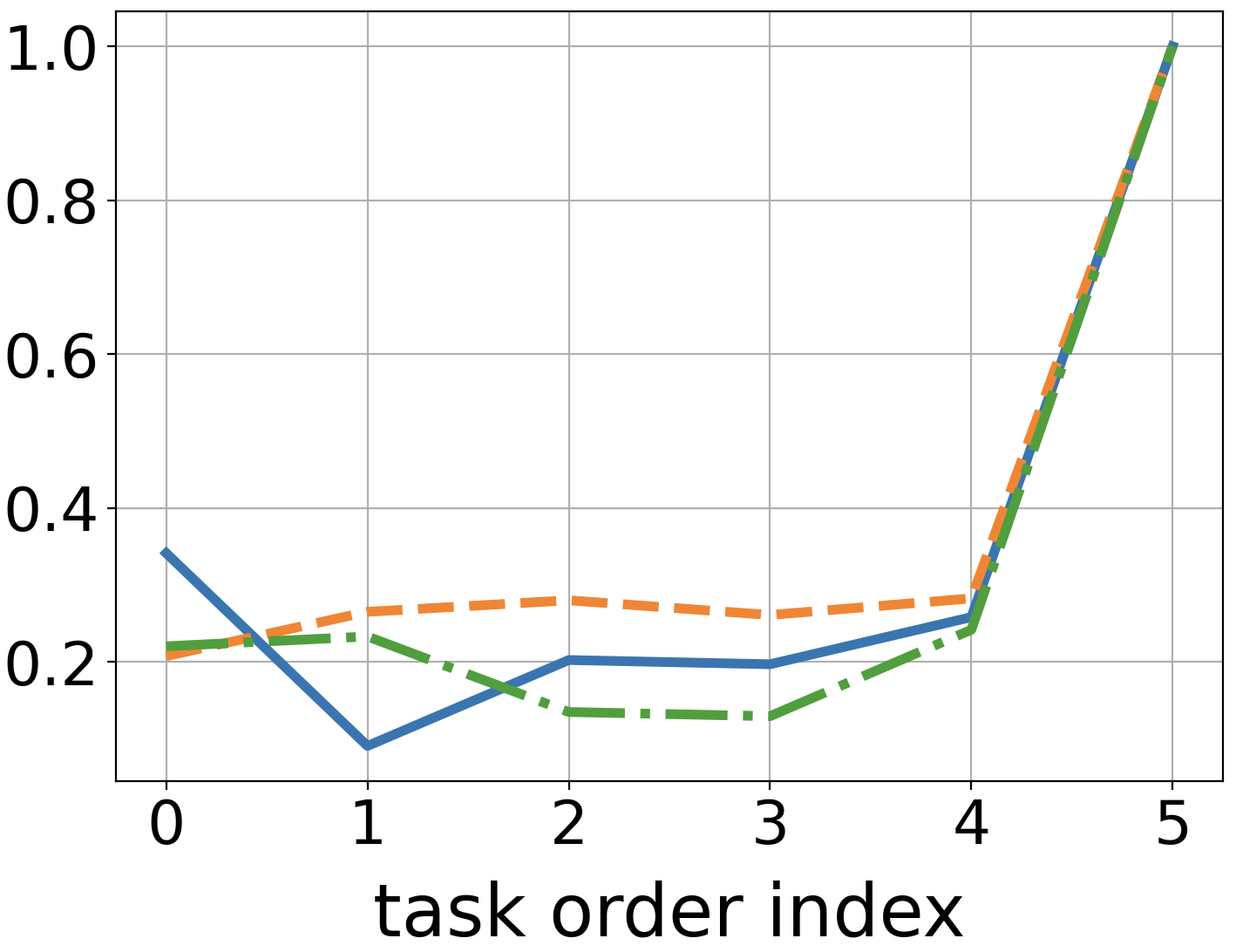}
    \label{fig:o1}}
    \subfigure[]{\includegraphics[width=0.21\textwidth]{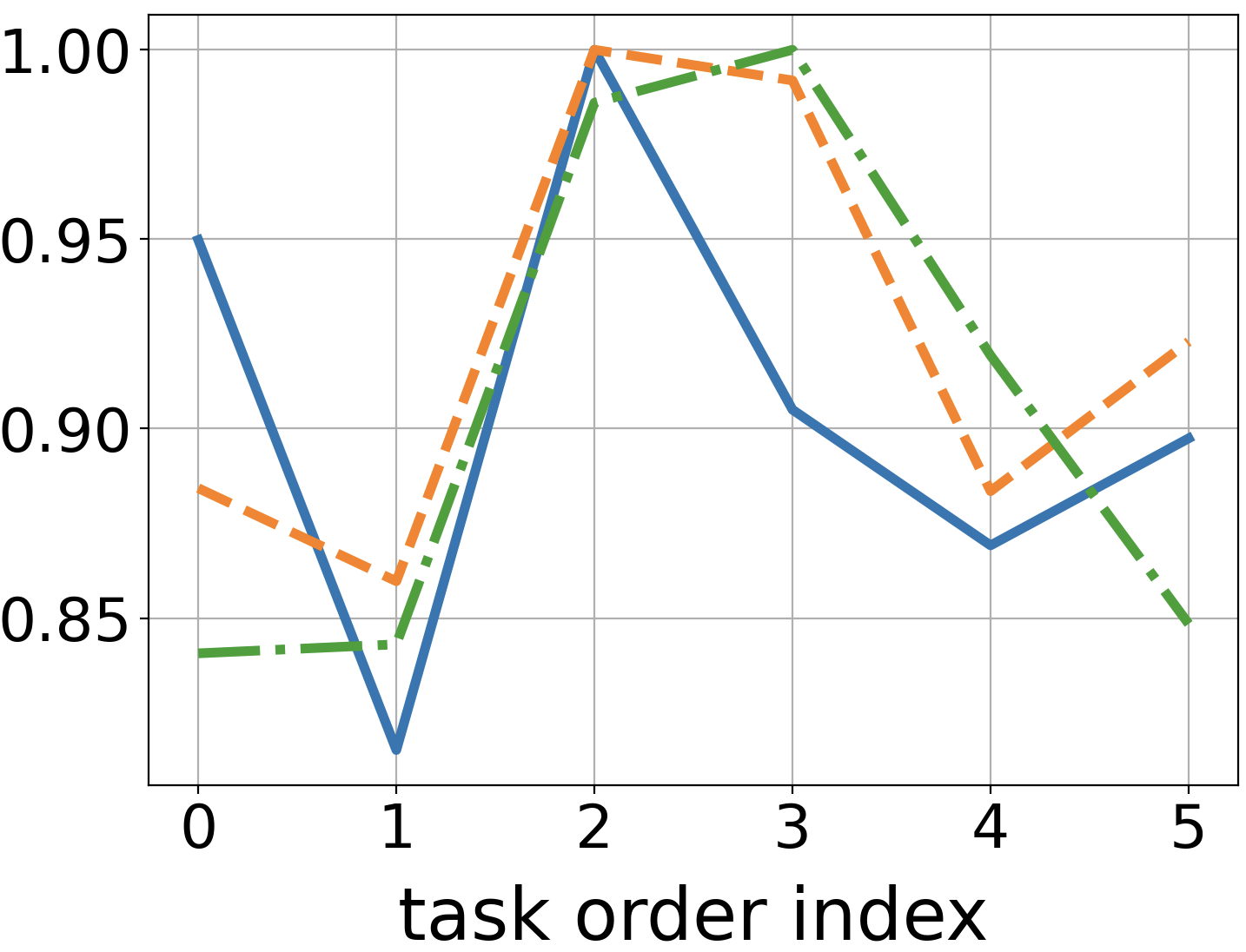}
    \label{fig:o2}}

    \caption{Impact of task similarity and task order. (a) When $T=2$, both forgetting and generalization error decrease when two tasks have more overlapping classes. (b) When $T=4$, the forgetting surprisingly increases when the first two tasks have more overlapping classes; `forgetting\_0', `forgetting\_1' and `forgetting\_2' correspond to three cases of the task setups (also in (c) and (d)). (c) Consider one special task and five same tasks in CL with $T=6$;  the task order index shows the order of the special task, and the smallest forgetting is achieved always when the special task is learnt in the first half for each case (we normalize the forgetting w.r.t. the worst forgetting in each case). (d) Consider two categories of tasks in CL with $T=4$; the task order indices $0$ and $1$ refer to the perfectly alternating orders, one of which always achieve the smallest forgetting among all possible orders. All the results are averaged over 3 random seeds.}
\label{fig:gnn}

\end{figure*}

(2) \emph{[Special case II: Equal Occurrence]~} There are two different categories ($C_1$ and $C_2$) of tasks, where tasks in the same category have the same optimal model; particularly, two categories contain the same number of tasks. If task $1\in C_1$ and task $2\in C_2$,  we will denote the task order as $(C_1, C_2)$. The following proposition characterizes the optimal task order in this case:
\begin{proposition}\label{pro:order2}
    Suppose $p\geq n+2$. For $T=4$ and $T=6$, the optimal task order to minimize forgetting is the perfectly alternating order, i.e., $(C_i,C_j,C_i,C_j)$ and $(C_i,C_j,C_i,C_j, C_i, C_j)$, where $i,j\in \{1, 2\}$ and $i\neq j$.
\end{proposition}
\cref{pro:order2} clearly shows that adjacent tasks always belong to different categories in the optimal task order, which leads to a more diverse task learning sequence. Intuitively, the alternating order maximizes the  memorization of each category by keeping practicing on different tasks. It can be further proved that the perfectly alternating order is also optimal for $T=6$ with three different categories (\cref{app:order}). Based on these results, we expect that such an alternating order may minimize forgetting for more general scenarios where the tasks contain multiple categories with equal cross-category task model distance.

The findings on the optimal task order indeed share similar insights with the surprising impact of task correlation on forgetting  mentioned earlier. Intuitively, learning more dissimilar tasks in the early stage facilitates the exploration of a larger feature space and expands the learnt feature space in CL, which can make the learning of similar tasks in the future much easier. In the meanwhile, the impact of task similarity among the early tasks continuously diminishes in CL with $T$ increasing, as suggested by the coefficients $c_{i,j}$ (which can be smaller for smaller $i$, $j$) in \cref{thm:exactform}. Therefore,  the negative impact of learning more dissimilar tasks on  forgetting is weaker when they are learnt in the early stage, compared to being learnt in the late stage.


\textbf{2) The optimal task ordering for minimizing forgetting and for minimizing generalization error are not always the same.}
Consider Special case I and Special case II. It can be shown that the optimal task orders for minimizing forgetting and generalization error are different in Special case I but same in Special case II. This would open up an interesting direction of finding the task order with balanced impact on forgetting and generalization error. A more detailed discussion can be found in \cref{app:order}.

\section{Implications on CL with DNN}


So far, we have explored different aspects that affect the performance of CL in overparameterized linear models.
More interestingly, we will show next that \cref{thm:exactform} can also shed light on CL in practice with DNNs, by reflecting on recent empirical observations and guiding improved designs therein.
More experimental details are in \cref{appen:exp}.

\subsection{Forgetting is not always monotonic with task similarity}
\label{subsec:correlation_dnn}

To see if our understandings about the impact of task similarity on forgetting can be carried over to CL with DNN, we conduct experiments on MNIST \cite{lecun1989handwritten} using a convolutional neural network to investigate the impact of task similarity therein. More specifically, we consider each task $i$ as a binary classification problem which seeks to decide if an image belongs to a task-specific label subset $Y_i$ of the classes, i.e., $Y_i\subset \{0,...,9\}$ in MNIST, and we control the task similarity through the degree of class overlapping between the task-specific subsets, e.g., task $i$ and $j$ are more similar if the cardinality of $Y_i\cap Y_j$ is larger.

We first consider the case with two tasks, where we fix $Y_1$ for task 1 as $\{0, 1, 2, 3, 4\}$ and change $Y_2$ for task 2 to have different numbers of overlapping classes with $Y_1$. As shown in \cref{fig:d2}, both forgetting and  generalization error decrease when the number of overlapping classes increases, i.e., the two tasks are more similar, which is indeed consistent with our analysis for the overparameterized linear models for $T=2$. More interestingly, this result also agrees with some recent studies \cite{ramasesh2020anatomy,lee2021continual,evron2022catastrophic}, which found that `intermediate task similarity' leads to the worst forgetting in a two-task setup using various notions of task similarity (different from our definition of task similarity using the optimal model gap), through either empirical studies or analyzing the upper bound of forgetting. We can build the connection based on the closed form of forgetting $F_2$ in \cref{eq:forget_two}.

Note that in \cref{eq:forget_two}
\begin{align*}
    \|\vw_2^*-\vw_1^*\|^2=\|\vw_{2}^*\|^2+\|\vw_{1}^*\|^2-2\langle \vw_{1}^*, \vw_{2}^*\rangle
\end{align*}
and we can divide the task correlation into three cases depending on the value of $\langle \vw_{1}^*, \vw_{2}^*\rangle$:
 (1) $\langle \vw_{1}^*, \vw_{2}^*\rangle=0$: Two tasks are orthogonal in the sense that they share no common features, i.e., $\gS_1\cap\gS_2=\emptyset$;
 (2) $\langle \vw_{1}^*, \vw_{2}^*\rangle>0$: Two tasks share some common features and are `positively' correlated;
 (3) $\langle \vw_{1}^*, \vw_{2}^*\rangle<0$: Two tasks share some common features but are `negatively' correlated. Compared to the first case when two tasks are orthogonal, it can be easily shown that  forgetting is worse when two tasks are negatively correlated even if they share some common features, which indeed corresponds to `the intermediate task similarity' in \cite{ramasesh2020anatomy,lee2021continual,evron2022catastrophic}. The reason behind is that in this case task 2 updates the model in the opposite direction to the model update of task 1, which inevitably leads to more forgetting in CL. Note that in \cref{fig:d2}, the non-overlapping case means that task 1 and 2 are negatively correlated because in this two-task case the image that is not in $Y_1$ must be in $Y_2$. On the other hand, the forgetting can even be negative when the two tasks are positively correlated.

We next consider the case with $T=4$, where we control the task similarity by changing $Y_2$ while fixing $Y_1$, $Y_3$ and $Y_4$. Here we let $(Y_1\cup Y_2) \cap (Y_3 \cup Y_4)=\emptyset$ as in \cref{subsec:correlation}. As shown in \cref{fig:d4},  forgetting surprisingly increases when task 1 and task 2 have more overlapping classes, which is also consistent with our analysis for the linear models. Indeed, this also justifies our observation that forgetting can decrease when the adjacent tasks are more dissimilar when studying the impact of task order.

\subsection{Diversify the tasks in the early stage and order dissimilar tasks adjacently}
\label{subsec:order_dnn}

We also evaluate the impact of task ordering on forgetting in CL with DNN, by constructing the tasks using a similar strategy as in 
\cref{subsec:correlation_dnn}. More specifically, we consider two different scenarios: (1) $T=6$, where the task sequence includes one special task and five same tasks; (2) $T=4$, where the task sequence includes two categories of tasks and each has two same tasks. 

\cref{fig:o1} demonstrates forgetting in the first scenario w.r.t. the learning order of the special task, and three plots correspond to three different cases, respectively. It is clear that for all three cases,
the optimal order of the special task to minimize forgetting is always in the first half of the sequence. 
For the second scenario, we evaluate  forgetting in \cref{fig:o2} for all six possible task orders, where task indices $0$ and $1$ refer to the perfectly alternating order. We can see that the smallest forgetting is also achieved in the perfectly alternating order. These results indicate that  our findings in \cref{subsec:order} for the overparameterized linear models can also be carried over to CL with DNN,
i.e., the optimal task order should diversify the tasks in the early stage and learn more different tasks adjacently. Such an implication is indeed consistent with the empirical observations in recent studies \cite{li2022provable,bell2022effect}.
Note that in both \cref{fig:o1} and \cref{fig:o2}, we normalize forgetting w.r.t. the worst forgetting in each case.

\subsection{Weight the fresher old tasks more in forward knowledge transfer}
\label{sec:trgp}

Recently, there has been increasing interest in CL on leveraging task correlation to facilitate  knowledge transfer \cite{ke2020continual,lin2022trgp,lin2022beyond}, which first selects the most correlated old tasks with the current task and then designs algorithms to directly increase the knowledge transfer between correlated tasks. By investigating knowledge transfer in the linear models, we show that improved algorithms can be motivated to achieve better knowledge transfer.

Given a task $t$ in CL, the forward knowledge transfer \cite{veniat2020efficient} in the linear model can be  defined as 
\begin{align}\label{eq:fwt}
    \mathbb{E}[\|\vw_t-\vw_t^*\|^2]-\mathbb{E}[\|\vw^r_t-\vw_t^*\|^2],
\end{align}
where $\vw^r_t$ is the learnt model of task $t$ by starting from a random model. Intuitively, \cref{eq:fwt} characterizes the gap in the testing performance between  $\vw_t$ learnt in CL and $\vw^r_t$ learnt from scratch, for which a positive value means that the accumulated knowledge in CL benefits the learning of the current task. As the second term in \cref{eq:fwt} is independent with CL, it
 suffices to analyze $\mathbb{E}[\|\vw_t-\vw_t^*\|^2]$ for understanding the forward knowledge transfer. Based on \cref{lem:gapevolution} (\cref{app.useful_lemmas}), we can obtain
\begin{align*}
    \mathbb{E}[\|\vw_t-\vw_t^*\|^2]
    =r^t \|\vw_t^*\|^2+ \sum_{i=1}^t \frac{n r^{t-i}}{p}\|\vw_i^*-\vw_t^*\|^2+\frac{p\sigma^2}{p-n-1}.
\end{align*}
While it is intuitive that better forward knowledge transfer can be achieved when $\|\vw_i^*-\vw_t^*\|^2$ is smaller for the current task $t$ and the old task $i$, the impact of different old tasks on the current task is non-uniform, in the sense that a more recent old task $i$ (i.e., $t-i$ is smaller) has a larger effect on the forward knowledge transfer to task $t$. This result implies that fresher old tasks should contribute more when designing algorithms to leverage correlated old tasks to facilitate better forward knowledge transfer.


To verify this insight, we consider the TRGP algorithm proposed in \cite{lin2022trgp}. Specifically, TRGP
first selects the most correlated old tasks with the current task 
and reuses their knowledge through a scaled weight projection to facilitate forward knowledge transfer, where all the selected old tasks are treated equivalently. We slightly modify TRGP by
assigning a larger weight to the selected old task that is more recent to the current task, named as TRGP+, and evaluate its performance on standard CL benchmarks (PMNIST \cite{lopez2017gradient} and Split CIFAR-100 \cite{krizhevsky2009learning}) and DNN architectures. As shown in \cref{tab:scale}, TRGP+ outperforms TRGP in both accuracy and forgetting. Assigning a larger weight to the more recent correlated old task not only improves the forward knowledge transfer, but also increases the backward knowledge transfer by forcing the learnt model of the current task to be closer to the model of those highly correlated old tasks. 

\begin{table}[!htbp]

\centering

\caption{The averaged final testing accuracy (ACC) and backward transfer (BWT: negative value of forgetting, larger is better) over all the tasks on different datasets.  }
\label{tab:scale}
\begin{tabular}{cccccc}
\toprule
\multicolumn{1}{c}{\multirow{2}{*}{Method}} & \multicolumn{2}{c}{PMNIST} &  & \multicolumn{2}{c}{Split CIFAR-100}\\ \cmidrule{2-3} \cmidrule{5-6} 
\multicolumn{1}{c}{} & ACC(\%) & BWT(\%)   &  & ACC(\%) & BWT(\%)    \\ \midrule
TRGP                  & 96.34   & -0.8 &  & 74.46   & -0.9   \\ \midrule
TRGP+                & 
\textbf{96.75}   & \textbf{-0.46} &  &    
\textbf{75.31}   & \textbf{0.13} \\ \bottomrule
\end{tabular}
\end{table}

\section{Conclusions}

In this work, we studied CL in the overparameterized linear models where each task is a linear regression problem and solved by using SGD. Under the assumption that each task has a sparse linear model with i.i.d. Gaussian features and noise, we derived the exact forms of both forgetting and generalization error, which built the key foundations of understanding the performance of CL. In particular, we investigated the impact of overparameterization,
task similarity and task ordering on both forgetting and generalization error. Experimental results on real datasets with DNNs indicated that our findings in linear models can even be carried over to CL in practice and leveraged to develop better algorithms.

\clearpage
\newpage

\bibliography{reference}
\bibliographystyle{plain}

\newpage
\appendix
\onecolumn

\section{Experimental Details}
\label{appen:exp}

\subsection{Experimental details for \cref{subsec:correlation_dnn} and \cref{subsec:order_dnn}}

\emph{Datasets.} We consider the MNIST dataset. For each task, we randomly select 200 samples for training and 1000 samples for testing. Different tasks have different subsets of classes.

\emph{DNN architecture and training details.} We use a five-layer neural network with two convolutional layers and three fully-connected layers. Relu is used for the first four layers and Sigmoid is used for the last layer. The first convolutional layer is followed  by 2D max-pooling operation with stride of 2. We learn each task by using SGD with a learning rate of $0.1$ for 600 epochs. The forgetting and overall generalization error are evaluated as in \cref{eq:forget_linear} and \cref{eq:generalization}, respectively, while here $\gL_t(\vw)$ is defined as the mean-squared test error instead of \cref{eq:def_model_error}.

\emph{Task setups.} For \cref{fig:d2}, we consider the following setup: 
\begin{itemize}
    \item task 1: $(0, 1, 2, 3, 4)$.
    \item task 2: $(5, 6, 7, 8, 9)$, $(4, 5, 6, 7, 8)$, $(3, 4, 5, 6, 7)$, $(2, 3, 4, 5, 6)$, $(1, 2, 3, 4, 5)$, $(0, 1, 2, 3, 4)$, which correspond to the different numbers of overlapping classes with task 1.
\end{itemize}

For \cref{fig:d4}, we randomly select three different setups:
\begin{itemize}
    \item `forgetting\_0': 
        \begin{itemize}
            \item task 1: $(0, 1, 2)$.
            \item task 2: $(3, 4, 5)$, $(2, 3, 4)$, $(1, 2, 3)$, $(0, 1, 2)$, which correspond to the different numbers of overlapping classes with task 1.
            \item task 3: $(7, 8, 9)$.
            \item task 4: $(7, 8 ,9)$.
            \end{itemize}
    \item `forgetting\_1': 
        \begin{itemize}
            \item task 1: $(3, 4, 5)$.
            \item task 2: $(0, 1, 2)$, $(1, 2, 3)$, $(2, 3, 4)$, $(3, 4, 5)$, which correspond to the different numbers of overlapping classes with task 1.
            \item task 3: $(6, 7, 8)$.
            \item task 4: $(7, 8 ,9)$.
            \end{itemize}
    \item `forgetting\_2': 
        \begin{itemize}
            \item task 1: $(0, 1, 2)$.
            \item task 2: $(7, 8 ,9)$, $(2, 7, 8)$, $(1, 2, 7)$, $(0, 1, 2)$, which correspond to the different numbers of overlapping classes with task 1.
            \item task 3: $(4, 5, 6)$.
            \item task 4: $(4, 5, 6)$.
            \end{itemize}
\end{itemize}

For \cref{fig:o1}, we randomly select three different setups:
\begin{itemize}
    \item `forgetting\_0': the special task is $(4, 5, 6, 7)$ and the other tasks are $(0, 1, 2, 3)$.
    \item `forgetting\_1': the special task is $(0, 1, 2, 3)$ and the other tasks are $(5, 6, 7, 8)$.
    \item `forgetting\_2': the special task is $(3, 4, 5, 6)$ and the other tasks are $(1, 2, 7, 8)$.
\end{itemize}

For \cref{fig:o2}, we randomly select three different setups:
\begin{itemize}
    \item `forgetting\_0': the two task categories are $(4, 5, 6, 7)$ and $(1, 2, 4, 5)$, and the task order indices are:
    \begin{itemize}
        \item `0': $(4, 5, 6, 7)$, $(1, 2, 4, 5)$, $(4, 5, 6, 7)$, $(1, 2, 4, 5)$.
        \item `1': $(1, 2, 4, 5)$, $(4, 5, 6, 7)$, $(1, 2, 4, 5)$, $(4, 5, 6, 7)$.
        \item `2': $(4, 5, 6, 7)$, $(4, 5, 6, 7)$, $(1, 2, 4, 5)$, $(1, 2, 4, 5)$.
        \item `3': $(1, 2, 4, 5)$, $(1, 2, 4, 5)$, $(4, 5, 6, 7)$, $(4, 5, 6, 7)$.
        \item `4': $(4, 5, 6, 7)$, $(1, 2, 4, 5)$, $(1, 2, 4, 5)$, $(4, 5, 6, 7)$.
        \item `5': $(1, 2, 4, 5)$, $(4, 5, 6, 7)$, $(4, 5, 6, 7)$, $(1, 2, 4, 5)$.
    \end{itemize}
    \item `forgetting\_1': the two task categories are $(4, 5, 6, 7)$ and $(2, 3, 4, 5)$, and the task order indices are:
    \begin{itemize}
        \item `0': $(4, 5, 6, 7)$, $(2, 3, 4, 5)$, $(4, 5, 6, 7)$, $(2, 3, 4, 5)$.
        \item `1': $(2, 3, 4, 5)$, $(4, 5, 6, 7)$, $(2, 3, 4, 5)$, $(4, 5, 6, 7)$.
        \item `2': $(4, 5, 6, 7)$, $(4, 5, 6, 7)$, $(2, 3, 4, 5)$, $(2, 3, 4, 5)$.
        \item `3': $(2, 3, 4, 5)$, $(2, 3, 4, 5)$, $(4, 5, 6, 7)$, $(4, 5, 6, 7)$.
        \item `4': $(4, 5, 6, 7)$, $(2, 3, 4, 5)$, $(2, 3, 4, 5)$, $(4, 5, 6, 7)$.
        \item `5': $(2, 3, 4, 5)$, $(4, 5, 6, 7)$, $(4, 5, 6, 7)$, $(2, 3, 4, 5)$.
    \end{itemize}
    \item `forgetting\_2': the two task categories are $(6, 7, 8, 9)$ and $(3, 4, 5, 6)$, and the task order indices are:
    \begin{itemize}
        \item `0': $(6, 7, 8, 9)$, $(3, 4, 5, 6)$, $(6, 7, 8, 9)$, $(3, 4, 5, 6)$.
        \item `1': $(3, 4, 5, 6)$, $(6, 7, 8, 9)$, $(3, 4, 5, 6)$, $(6, 7, 8, 9)$.
        \item `2': $(6, 7, 8, 9)$, $(6, 7, 8, 9)$, $(3, 4, 5, 6)$, $(3, 4, 5, 6)$.
        \item `3': $(3, 4, 5, 6)$, $(3, 4, 5, 6)$, $(6, 7, 8, 9)$, $(6, 7, 8, 9)$.
        \item `4': $(6, 7, 8, 9)$, $(3, 4, 5, 6)$, $(3, 4, 5, 6)$, $(6, 7, 8, 9)$.
        \item `5': $(3, 4, 5, 6)$, $(6, 7, 8, 9)$, $(6, 7, 8, 9)$, $(3, 4, 5, 6)$.
    \end{itemize}
\end{itemize}

\subsection{Experimental details for \cref{sec:trgp}}

\subsubsection{TRGP vs TRGP+}
TRGP \cite{lin2022trgp} seeks to solve the following optimization problem for the current task $t$:
\begin{align}\label{eq:trgp}
    \smash{\min_{\{\vw^l\}_l, \{\mQ_{j,t}^l\}_{l,j\in\gtr_t^l}}}~~& \gL(\{\vw^l_{eff}\}_l, \gD_t),\nonumber\\
    s.t~~~~~~~~~~~~~~~& \vw^l_{eff}=\vw^l+\sum\nolimits_{j\in\gtr_t^l} [\proj^Q_{S_j^l}(\vw^l)-\proj_{S_j^l}(\vw^l)],
\end{align}
where $\vw^l$ is the DNN weight for the layer $l$, and $S_j^l$ denotes  the input subspace of the layer $l$ for the old task $j<t$, which can be constructed by using SVD on the representation matrix for that layer. Two important designs are introduced in \cref{eq:trgp}: 
\begin{itemize}
    \item The trust region $\gtr_t^l$: $\gtr_t^l$ denotes the set of the most correlated old tasks selected for task $t$ based on some correlation evaluation metric in a layer-wise manner. The purpose here is to select the most correlated old tasks and facilitate the forward knowledge transfer by reusing the learnt knowledge of the old tasks in $\gtr_t^l$.

    \item The scaled weight projection $\proj^Q_{S_j^l}(\vw^l)$: $\proj^Q_{S_j^l}(\vw^l)$ is developed to reuse the learnt model of the selected old tasks in $\gtr_t^l$. Specifically, for any $j\in \gtr_t^l$, 
    \begin{align*}
        \proj^Q_{S_j^l}(\vw^l)=\vw_{t-1}^l\mB_j^l \mQ_{j,t}^l (\mB_j^l)'
    \end{align*}
  where $\mB_j^l$ is the bases matrix for the subspace $S_j^l$, and $\mQ_{j,t}^l$ is the scaling matrix to scale the weight projection onto $S_j^l$. In contrast, $\proj_{S_j^l}(\vw^l)=\vw_{t-1}^l\mB_j^l  (\mB_j^l)'$ is the standard weight projection onto $S_j^l$. Since the learnt knowledge for the old task $j$ is indeed $\proj_{S_j^l}(\vw^l)$, scaling the projection provides a way to reuse this knowledge directly for learning the task $t$. Intuitively, $\proj^Q_{S_j^l}(\vw^l)-\proj_{S_j^l}(\vw^l)$ characterizes the boosted forward knowledge transfer from old task $j\in \gtr_t^l$ to task $t$.
\end{itemize}

However, as shown in \cref{eq:trgp}, all the selected old tasks in $\gtr_t^l$ are treated equivalently in the effective weight $\vw^l_{eff}$, which could be suboptimal. As suggested by our theoretical results, we proposed a slightly modified version of TRGP, i.e., TRGP+, by assigning  non-uniform weights for the most correlated old tasks selected in $\gtr_t^l$:
\begin{align}\label{eq:trgp+}
    \smash{\min_{\{\vw^l\}_l, \{\mQ_{j,t}^l\}_{l,j\in\gtr_t^l}}}~~& \gL(\{\vw^l_{eff}\}_l, \gD_t),\nonumber\\
    s.t~~~~~~~~~~~~~~~& \vw^l_{eff}=\vw^l+\sum\nolimits_{j\in\gtr_t^l} \lambda_j [\proj^Q_{S_j^l}(\vw^l)-\proj_{S_j^l}(\vw^l)],
\end{align}
where $\lambda_j>\lambda_{j'}$ if $t-j<t-j'$ for both $j$, $j'\in \gtr_t^l$.

\subsubsection{Experimental setup}

\emph{Datasets.} We consider two standard benchmarks in CL: (1) \textbf{PMNIST}: 10 sequential tasks will be created using different permutations, where each task has 10 classes; (2) \textbf{Split CIFAR-100}: The entire dataset of CIFAR-100 will be splitted into 10 group, where each task is a 10-way multi-class classification problem for each group.

\emph{DNN architecture and training details.} Following \cite{lin2022trgp},  we use a 3-layer fully-connected network with 2 hidden layer of 100 units for PMNIST, and train the network for 5 epochs  with a batch size of 10 for each task. For Split CIFAR-100, we use a version of 5-layer AlexNet, and train the network for a maximum of 200 epochs with early stopping for each task. Two most correlated old tasks are selected for the current task for each layer, and we assign a larger weight of $1.2$ to the more recent old task and $0.8$ to the other one.

\emph{Evaluation metrics.} The performance is evaluated based on ACC, the average final accuracy over all tasks, and Backward Transfer (BWT) which measures the forgetting of old tasks when learning new tasks. Specfically, ACC and BWT are defined as:
{\small
\begin{align}\label{eq:metric}
    ACC = \frac{1}{T}\sum\nolimits_{i=1}^{T} A_{T,i}, 
    BWT = \frac{1}{T-1}\sum\nolimits_{i=1}^{T-1} A_{T,i} - A_{i,i}
\end{align}
}%
where $A_{T,i}$ is the accuracy of the model on $i$-th task after learning the $T$-th task sequentially.

\section{Useful Lemmas}\label{app.useful_lemmas}
The following lemma characterizes the solution to the optimization problem \cref{eq:op} for task $t$:
\begin{lemma}\label{lem:solution}
    The solution to the optimization problem \cref{eq:op}, i.e., the learnt model for task $t$, is given by
    \begin{align}\label{eq:solution}
        \vw_t=\vw_{t-1}+\mX_t(\mX_t^{\top}\mX_t)^{-1}\left(\vy_t-\mX_t^{\top}\vw_{t-1}\right).
    \end{align}
\end{lemma}
In the overparameterized case, multiple $\vw_t$ exist to perfectly fit $(\mX_t)^{\top}\vw=\vy_t$, and solving \cref{eq:op} picks the one that has minimum $l^2$ distance to $\vw_{t-1}$. Therefore, the solution in \cref{eq:solution} not only incorporates the information of current task $t$ through $\gD_t$ but also depends on the previous model evolution trajectory in CL. 

By leveraging the recent advance in \cite{belkin2020two}, we can have the following lemma about the evolution of $\mathbb{E}[\|\vw_t-\vw_i^*\|^2]$:
\begin{lemma}\label{lem:gapevolution}
 Suppose $p\geq n+2$. For any task $t\in [1, T-1]$ and any old task $i\in [1, t]$, the following equation holds:
 {\small
 \begin{align*}
     &\mathbb{E}[\|\vw_{t+1}-\vw_i^*\|^2]\\
     =&\left(1-\frac{n}{p}\right)\E[\|\vw_t-\vw_i^*\|^2]+\frac{n}{p}\|\vw_{t+1}^*-\vw_i^*\|^2+\frac{n\sigma^2}{p-n-1}.
 \end{align*}
 }%
\end{lemma}

\section{Additional Results}

\subsection{Characterization of negative forgetting}

As shown in \cref{fig:d2}, the forgetting can even be negative when the two tasks are positively correlated. Intuitively, because the common features play a similar role in these two tasks, task 2 updates the model in a favorable direction for task 1, which could even result in better performance of task 1 due to the backward knowledge transfer herein. A formal quantification of the condition for better performance of task 1 can be found in the following proposition:

\begin{proposition}\label{pro:two_task_back}
  Suppose $\sigma^2<\frac{p-n-1}{p}\|\vw_1^*\|^2$ and $p\geq n+2$. The learning of task 2 would lead to a better model for task 1, i.e., $\E[F_2]\leq 0$, if
  \begin{align*}
      2\langle \vw_{1,\gS_1}^*, \vw_{2,\gS_2}^*\rangle\geq \frac{n}{p}\|\vw_1^*\|^2+\|\vw_2^*\|^2+\frac{(p-n)\sigma^2}{p-n-1}.
  \end{align*}
\end{proposition}

\subsection{Evolution of forgetting}

We can also characterize the evolution of forgetting after learning new tasks. 
Based on the definition of forgetting, we have
\allowdisplaybreaks
\begin{align*}
    \E[F_t]&=\frac{1}{t-1}\sum_{i=1}^{t-1} \E\left[\|\vw_t-\vw_i^*\|^2-\|\vw_i-\vw_i^*\|^2\right],\\
    \E[F_{t+1}]&=\frac{1}{t}\sum_{i=1}^{t} \E\left[\|\vw_{t+1}-\vw_i^*\|^2-\|\vw_i-\vw_i^*\|^2\right].
\end{align*}
Rearranging the above equations gives
\begin{align*}
    \sum_{i=1}^{t-1} \E[\|\vw_t-\vw_i^*\|^2]&= (t-1)\E[F_t]+\sum_{i=1}^{t-1} \E[\|\vw_i-\vw_i^*\|^2],\\
    \sum_{i=1}^{t-1} \E[\|\vw_{t+1}-\vw_i^*\|^2]&=t \E[F_{t+1}]+\sum_{i=1}^{t} \E[\|\vw_i-\vw_i^*\|^2]-\E[\|\vw_{t+1}-\vw_t^*\|^2].
\end{align*}

Based on the relationship between $\E[\|\vw_t-\vw_i^*\|^2]$ and $\E[\|\vw_{t+1}-\vw_i^*\|^2]$ characterized in \cref{lem:gapevolution}, it can be seen that
\begingroup
\allowdisplaybreaks
\begin{align*}
    &\sum_{i=1}^{t-1} \E[\|\vw_{t+1}-\vw_i^*\|^2]\\
    =&t \E[F_{t+1}]+\sum_{i=1}^{t} \E[\|\vw_i-\vw_i^*\|^2]-\E[\|\vw_{t+1}-\vw_t^*\|^2]\\
    =&\sum_{i=1}^{t-1}\left\{\left(1-\frac{n}{p}\right)\E[\|\vw_t-\vw_i^*\|^2]+\frac{n}{p}\|\vw_{t+1}^*-\vw_i^*\|^2+\frac{n\sigma^2}{p-n-1}\right\}\\
    =&\left(1-\frac{n}{p}\right)\sum_{i=1}^{t-1} \E[\|\vw_t-\vw_i^*\|^2]+\frac{n}{p}\sum_{i=1}^{t-1}\|\vw_{t+1}^*-\vw_i^*\|^2+\frac{n\sigma^2(t-1)}{p-n-1}\\
    =& \left(1-\frac{n}{p}\right)\left\{(t-1)\E[F_t]+\sum_{i=1}^{t-1} \E[\|\vw_i-\vw_i^*\|^2]\right\}
    +\frac{n}{p}\sum_{i=1}^{t-1}\|\vw_{t+1}^*-\vw_i^*\|^2+\frac{n\sigma^2(t-1)}{p-n-1},
\end{align*}
\endgroup
such that
\begin{align}\label{eq:re_1}
    t \E[F_{t+1}]=& (t-1)\left(1-\frac{n}{p}\right)\E[F_t]+\left(1-\frac{n}{p}\right)\sum_{i=1}^{t-1} \E[\|\vw_i-\vw_i^*\|^2]+\frac{n}{p}\sum_{i=1}^{t-1}\|\vw_{t+1}^*-\vw_i^*\|^2\nonumber\\
    &+\frac{n\sigma^2(t-1)}{p-n-1}-\sum_{i=1}^{t} \E[\|\vw_i-\vw_i^*\|^2]+\E[\|\vw_{t+1}-\vw_t^*\|^2]\nonumber\\
    =& (t-1)\left(1-\frac{n}{p}\right)\E[F_t]-\frac{n}{p}\sum_{i=1}^{t-1} \E[\|\vw_i-\vw_i^*\|^2]-\E[\|\vw_t-\vw_t^*\|^2]\nonumber\\
    &+\frac{n}{p}\sum_{i=1}^{t-1}\|\vw_{t+1}^*-\vw_i^*\|^2+\E[\|\vw_{t+1}-\vw_t^*\|^2]+\frac{n\sigma^2(t-1)}{p-n-1}.
\end{align}

Let $i=t$ in \cref{lem:gapevolution}. We can show that
\begin{align}\label{eq: nbh}
    &\E[\|\vw_{t+1}-\vw_t^*\|^2-\|\vw_t-\vw_t^*\|^2]\nonumber\\
    =&\frac{n}{p}\|\vw_{t+1}^*-\vw_t^*\|^2-\frac{n}{p}\E[\|\vw_t-\vw_t^*\|^2]+\frac{n\sigma^2}{p-n-1}.
\end{align}

By substituting \cref{eq: nbh} back to \cref{eq:re_1}, we can have
\begin{align}\label{eq:forget_another}
    \E[F_{t+1}]
    =&\frac{t-1}{t}\left(1-\frac{n}{p}\right)\E[F_t]+\frac{n}{tp}\sum_{i=1}^{t-1} \left\{\|\vw_{t+1}^*-\vw_i^*\|^2-\E[\|\vw_i-\vw_i^*\|^2]\right\}\nonumber\\
    &+\frac{n}{tp}\left\{\|\vw_{t+1}^*-\vw_t^*\|^2-\E[\|\vw_t-\vw_t^*\|^2]\right\}+\frac{n\sigma^2}{p-n-1}\nonumber\\
    =&\frac{t-1}{t}\left(1-\frac{n}{p}\right)\E[F_t]+\frac{n}{tp}\sum_{i=1}^{t} \left\{\|\vw_{t+1}^*-\vw_i^*\|^2-\E[\|\vw_i-\vw_i^*\|^2]\right\}+\frac{n\sigma^2}{p-n-1}.
\end{align}

\subsection{Impact of overparameterization}
\label{app:over}

\textbf{1) Forgetting approaches zero with more parameters.} In \cref{eq:forget_close}, when $p\to \infty$, we have $r\to 1$, which implies that $(r^T-r^i)\to 0$ and $c_{i,j}\to 0$. Therefore, we can conclude that $\mathbb{E}[F_T]\to 0$ when $p\to \infty$. An intuitive explanation is that with more parameters, the model has a larger ``memory'' such that it can remember all knowledge of previous tasks, i.e., zero forgetting.

\textbf{2) More parameters can alleviate the negative impact of task dissimilarity on generalization error.} Term~G2 in \cref{eq:general_close} describes the effect of task dissimilarity on $G_T$. When $p\to \infty$, Term~G2 approaches zero, which indicates that the negative impact of task dissimilarity on generalization error diminishes. In some special cases, we can further show that Term~G2 is monotonically decreasing with respect to $p$, e.g., $T=2$ shown in \cref{eq:generalization_two}. A more general\footnote{For general $T$, this requirement holds if the ground truth of each task has the same power and is orthogonal to each other, i.e., $\|\vw_i^*\|^2=\|\vw_j^*\|^2$ and $(\vw_i^*)^T\vw_j^*=0$ for all $i\neq j$.} case is when $\sum_{k=1}^T\|\vw_k^*-\vw_i^*\|^2=C$ for all task $i$, we have Term~G2 $=\frac{1-r^T}{T}C$ which is also monotonically decreasing w.r.t. $p$.

\subsection{Impact of task order}
\label{app:order}

(1) \textbf{[Special case III]~} There are three categories ($C_1$, $C_2$ and $C_3$) of tasks: each category contains the same number of tasks; the tasks are same in the same category but different across categories. Without loss of generality,  we assume that for any task $i$ and $j$
\begin{align*}
     \|\vw_i^*-\vw_j^*\|^2=
     \begin{cases}
        0, & \text{if  $i, j\in C_m$  for $m\in\{1,2,3\}$;}\\
        1, & \text{else.}
     \end{cases}
\end{align*}
Based on \cref{thm:exactform}, we
can show that the optimal task order for Special case III follows a similar structure of that for Special case II, 
as characterized in the following proposition:
\begin{proposition}\label{pro:order3}
    Suppose $p\geq n+2$. For $T=6$, the optimal task order to minimize forgetting is the perfectly alternating order, i.e., $(C_i,C_j,C_k,C_i, C_j, C_k)$, where $i,j,k\in \{1, 2, 3\}$, $i\neq j$, $i\neq k$ and $j\neq k$.
\end{proposition}

(2) \textbf{[The optimal task order can be different for minimizing  forgetting and  generalization error]}
\label{appen:order_compare}

\emph{[Special case I]~} As shown in \cref{pro:order1}, the optimal task order to minimize forgetting is to learn the special task between the $2nd$ place and the $\frac{T}{2}th$ place. In stark  contrast, this special task, which has the largest value of $\sum_{k=1}^T\|\vw_k^*-\vw_i^*\|^2$, should be always learnt in the very first place in order to minimize the generalization error, i.e., $i=1$. The underlying rationale is that  the generalization error characterizes the average testing performance of
the final model on all tasks, which
is maximized when the final model works the best for the majority. Therefore, in this case the optimal order for minimizing forgetting is different from that for minimizing  generalization error.

\emph{[Special case II]~} As shown in \cref{pro:order2}, the optimal task order to minimize forgetting is the perfectly alternating order. In contrast, the task order indeed does not affect the generalization performance, because $\sum_{k=1}^T\|\vw_k^*-\vw_i^*\|^2$
is same for every task $i\in\mathbb{T}$. In this case, the optimal task order for minimizing  forgetting is also `optimal’ for minimizing
generalization error. That is to say, we can find an optimal task order to minimize 
forgetting and   generalization error simultaneously.

\section{Proofs}

\subsection{Proof of \cref{lem:solution}}

Let $\hat{\vw}=\vw-\vw_{t-1}$. It is clear that \cref{eq:op} can be reformulated as 
\begin{align}\label{eq:ref}
    \min&~~ \|\hat{\vw}\|_2,\\
    s.t.&~~ \mX_t^{\top}\hat{\vw}=\vy_t-\mX_t^{\top} \vw_{t-1}.\nonumber
\end{align}
For the overparameterized case, $\mX_t^{\top}\mX_t$ is invertible.
Using the Lagrange multipliers, we can get
\begin{align*}
    \min_{\hat{\vw},\lambda}~~\frac{\hat{\vw}^{\top} \hat{\vw}}{2}+\lambda^T[\mX_t^{\top}\hat{\vw}-(\vy_t-\mX_t^{\top} \vw_{t-1})].
\end{align*}
By setting the derivative w.r.t. $\hat{\vw}$ to 0, it follows that
\begin{align}\label{eq:d0}
    \hat{\vw}^*=- \mX_t\lambda
\end{align}
such that
\begin{align*}
    \mX_t^{\top} \hat{\vw}^* = -\mX_t^{\top}\mX_t \lambda = \vy_t-\mX_t^{\top} \vw_{t-1}.
\end{align*}

Therefore, 
\begin{align}\label{eq:lambda}
    \lambda=-(\mX_t^{\top}\mX_t)^{-1}(\vy_t-\mX_t^{\top} \vw_{t-1}).
\end{align}

By substituting \cref{eq:lambda} into \cref{eq:d0}, we can have
\begin{align*}
    \hat{\vw}^*=\mX_t (\mX_t^{\top}\mX_t)^{-1}(\vy_t-\mX_t^{\top} \vw_{t-1})
\end{align*}
such that
\begin{align*}
    \vw_t=\vw_{t-1}+\mX_t (\mX_t^{\top}\mX_t)^{-1}(\vy_t-\mX_t^{\top} \vw_{t-1}).
\end{align*}

\subsection{Proof of \cref{lem:gapevolution}}
Let $\mP_t\defeq \mX_t(\mX_t^{\top}\mX_t)^{-1}\mX_t^{\top}$ and $\mX_t^\dagger\defeq \mX_t(\mX_t^\top \mX_t)^{-1}$ for any $t\in\mathbb{T}$, where $\mP_t$ characterizes the projection onto the row space of $\mX_t^{\top}$.  Based on \cref{lem:solution}, we have
\begin{align}\label{eq:evol}
    \vw_{t+1}= (\mI-\mP_{t+1})\vw_t + \mP_{t+1}\vw_{t+1}^*+\mX_{t+1}^{\dagger}\vz_{t+1}.
\end{align}
Intuitively, the learnt model $\vw_{t+1}$ for task $t+1$ is an `interpolation' between the learnt model $\vw_t$ for task $t$ and the optimal task model $\vw_{t+1}^*$ for task $t+1$, while being perturbed by the random noise $z_{t+1}$.

Let $H=(\mI-\mP_{t+1})(\vw_t-\vw_i^*) + \mP_{t+1}(\vw_{t+1}^*-\vw_i^*)$.
Based on \cref{eq:evol}, we can know that
\allowdisplaybreaks
\begin{align}\label{eq:wo_e}
    &\E[\|\vw_{t+1}-\vw_i^*\|^2]\nonumber\\
    =&\E[\|(\mI-\mP_{t+1})\vw_t + \mP_{t+1}\vw_{t+1}^*+\mX_{t+1}^{\dagger}\vz_{t+1}-\vw_i^*\|^2]\nonumber\\
    =&\E[\|(\mI-\mP_{t+1})(\vw_t-\vw_i^*) + \mP_{t+1}(\vw_{t+1}^*-\vw_i^*)+\mX_{t+1}^{\dagger}\vz_{t+1}\|^2]\nonumber\\
    =& \E[\|H+\mX_{t+1}^{\dagger}\vz_{t+1}\|^2]\nonumber\\
    =& \underbrace{\E[\|H\|^2]}_{(a)}+\underbrace{2\E[\langle H, \mX_{t+1}^{\dagger}\vz_{t+1}\rangle]}_{(b)}+\underbrace{\E[\|\mX_{t+1}^{\dagger}\vz_{t+1}\|^2]}_{(c)}.
\end{align}

(1) For the term (a), we have
\begin{align}\label{eq:a1}
    \E[\|H\|^2]=&\E[\|(\mI-\mP_{t+1})(\vw_t-\vw_i^*) + \mP_{t+1}(\vw_{t+1}^*-\vw_i^*)\|^2]\nonumber\\
    =& \E[\|(\mI-\mP_{t+1})(\vw_t-\vw_i^*)\|^2]+\E[\|\mP_{t+1}(\vw_{t+1}^*-\vw_i^*)\|^2]+2\E[\langle (\mI-\mP_{t+1})(\vw_t-\vw_i^*), \mP_{t+1}(\vw_{t+1}^*-\vw_i^*)\rangle]\nonumber\\
    \overset{(a)}{=}& \E[\|(\mI-\mP_{t+1})(\vw_t-\vw_i^*)\|^2]+\E[\|\mP_{t+1}(\vw_{t+1}^*-\vw_i^*)\|^2]\nonumber\\
    \overset{(b)}{=}& \E[\|\vw_t-\vw_i^*\|^2]-\E[\|\mP_{t+1}(\vw_t-\vw_i^*)\|^2]+\E[\|\mP_{t+1}(\vw_{t+1}^*-\vw_i^*)\|^2]
\end{align}
where (a) is because of the orthogonality between $\mI-\mP_{t+1}$ and $\mP_{t+1}$, and (b) is due to the Pythagorean theorem.

Because $\mP_{t+1}$ is the orthogonal projection matrix for the row space of $\mX_{t+1}$, based on the rotational symmetry of the standard normal distribution, it follows that
\begin{align}\label{eq:p1}
    \E[\|\mP_{t+1}(\vw_{t+1}^*-\vw_i^*)\|^2]=\frac{n}{p}\|\vw_{t+1}^*-\vw_i^*\|^2,
\end{align}
and
\begin{align}\label{eq:p2}
    \E[\|\mP_{t+1}(\vw_t-\vw_i^*)\|^2]=\frac{n}{p}\E[\|\vw_t-\vw_i^*\|^2],
\end{align}
since $\mP_{t+1}$ is independent with $\vw_t$.

By substituting \cref{eq:p1} and \cref{eq:p2} back to \cref{eq:a1}, we can obtain that
\begin{align}\label{eq:a2}
    \E[\|H\|^2]=\left(1-\frac{n}{p}\right)\E[\|\vw_t-\vw_i^*\|^2]+\frac{n}{p}\|\vw_{t+1}^*-\vw_i^*\|^2.
\end{align}

(2) For the term (b), we have
\begin{align*}
    \E[\langle H, \mX_{t+1}^{\dagger} \vz_{t+1} \rangle]=&\E[\langle (\mI-\mP_{t+1})(\vw_t-\vw_i^*) + \mP_{t+1}(\vw_{t+1}^*-\vw_i^*), \mX_{t+1}^{\dagger} \vz_{t+1} \rangle]\\
    =& \E[\langle (\mI-\mP_{t+1})(\vw_t-\vw_i^*), \mX_{t+1}^{\dagger} \vz_{t+1} \rangle] + \E[\langle \mP_{t+1}(\vw_{t+1}^*-\vw_i^*), \mX_{t+1}^{\dagger} \vz_{t+1} \rangle].
\end{align*}
Because $(\mI-\mP_{t+1})$ is the projection onto the null space of $\mX_{t+1}^{\top}$ and $\mX_{t+1}^{\dagger} \vz_{t+1}$ is a vector in the row space of $\mX_{t+1}^{\top}$, it follows that
\begin{align}\label{eq:proj}
    \E[\langle (\mI-\mP_{t+1})(\vw_t-\vw_i^*), \mX_{t+1}^{\dagger} \vz_{t+1} \rangle]=0.
\end{align}

And since
\begin{align*}
     \E[\langle \mP_{t+1}(\vw_{t+1}^*-\vw_i^*), \mX_{t+1}^{\dagger} \vz_{t+1} \rangle] = \E[\langle (\mX_{t+1}^{\dagger})^{\top}\mP_{t+1}(\vw_{t+1}^*-\vw_i^*),  \vz_{t+1} \rangle]=0.
\end{align*}
we can know that
\begin{align}\label{eq:b}
    \E[\langle H, \mX_{t+1}^{\dagger} \vz_{t+1} \rangle]=0.
\end{align}

(3) For the term (c), we apply the ``trace trick" by following \cite{belkin2020two}. Specifically,
it can be first seen that
\begin{align*}    \|\mX_{t+1}^{\dagger}\vz_{t+1}\|^2=&\|\mX_{t+1}(\mX_{t+1}^\top \mX_{t+1})^{-1}\vz_{t+1}\|^2\\
=& tr((\mX_{t+1}^\top \mX_{t+1})^{-1}(\mX_{t+1}^\top \mX_{t+1})(\mX_{t+1}^\top \mX_{t+1})^{-1}\vz_{t+1}\vz_{t+1}^{\top})\\
=& tr((\mX_{t+1}^\top \mX_{t+1})^{-1}\vz_{t+1}\vz_{t+1}^{\top})
\end{align*}
Due to the independence between $\mX_{t+1}$ and the random noise $\vz_{t+1}$, we can have that
\begin{align*}
    \E[\|\mX_{t+1}^{\dagger}\vz_{t+1}\|^2]=&\E[tr((\mX_{t+1}^\top \mX_{t+1})^{-1}\vz_{t+1}\vz_{t+1}^{\top}))]\\
    =& tr[\E[(\mX_{t+1}^\top \mX_{t+1})^{-1}\vz_{t+1}\vz_{t+1}^{\top}]]\\
    =& tr(\E[(\mX_{t+1}^\top \mX_{t+1})^{-1}]\E[\vz_{t+1}\vz_{t+1}^{\top}])\\
    =& \sigma^2 tr(\E[(\mX_{t+1}^\top \mX_{t+1})^{-1}]).
\end{align*}
Since $(\mX_{t+1}^\top \mX_{t+1})^{-1}$ follows the inverse-Wishart distribution with identity scale matrix $\mI\in\sR^{n\times n}$ and $p$ degrees-of-freedom, and each diagonal entry of $(\mX_{t+1}^\top \mX_{t+1})^{-1}$ has a reciprocal that follows the $\chi^2$ distribution with $p-n+1$ degrees-of-freedom. Therefore, for $p\geq n+2$, 
\begin{align*}
    tr(\E[(\mX_{t+1}^\top \mX_{t+1})^{-1}])=\frac{n}{p-n+1},
\end{align*}
such that
\begin{align}\label{eq:c}
    \E[\|\mX_{t+1}^{\dagger}\vz_{t+1}\|^2]=\frac{n\sigma^2}{p-n+1}.
\end{align}

\cref{lem:gapevolution} can be proved by substituting \cref{eq:a2}, \cref{eq:b} and \cref{eq:c} to \cref{eq:wo_e}.

\subsection{Proof of \cref{thm:exactform}}\label{proof:thm:exactform}

Based on \cref{lem:gapevolution}, we can have that
\allowdisplaybreaks
\begin{align}\label{eq:first}
    \E[\|\vw_t-\vw_i^*\|^2]=&\left(1-\frac{n}{p}\right)^{t} \|\vw_0-\vw_i^*\|^2+\sum_{k=1}^{t} \left(1-\frac{n}{p}\right)^{t-k} \frac{n}{p}\|\vw_{k}^*-\vw_i^*\|^2\nonumber\\
    &+ \frac{n\sigma^2}{p-n-1}\sum_{k=1}^{t} \left(1-\frac{n}{p}\right)^{t-k}\nonumber\\
    =&\left(1-\frac{n}{p}\right)^{t} \|\vw_i^*\|^2+\sum_{k=1}^{t} \left(1-\frac{n}{p}\right)^{t-k} \frac{n}{p}\|\vw_{k}^*-\vw_i^*\|^2\nonumber\\
    &+ \frac{n\sigma^2}{p-n-1}\sum_{k=1}^{t} \left(1-\frac{n}{p}\right)^{t-k}\text{ (since $\vw_0=\bm{0}$)}.
\end{align}

Let $t=i$ in \cref{eq:first}. We have
\begin{align}\label{eq:second}
    \E[\|\vw_i-\vw_i^*\|^2]=&\left(1-\frac{n}{p}\right)^{i} \|\vw_i^*\|^2+\sum_{k=1}^{i} \left(1-\frac{n}{p}\right)^{i-k} \frac{n}{p}\|\vw_{k}^*-\vw_i^*\|^2\nonumber\\
    &+ \frac{n\sigma^2}{p-n-1}\sum_{k=1}^{i} \left(1-\frac{n}{p}\right)^{i-k}.
\end{align}

Based on \cref{eq:first} and \cref{eq:second}, we can obtain the closed form of $\E[F_T]$:
\begin{align*}
    &\E[F_T]\\
    =&\frac{1}{T-1}\sum_{i=1}^{T-1} \E\left[\|\vw_T-\vw_i^*\|^2-\|\vw_i-\vw_i^*\|^2\right]\\
    =&\frac{1}{T-1}\sum_{i=1}^{T-1} \Bigg\{\left(1-\frac{n}{p}\right)^{T} \|\vw_i^*\|^2+\sum_{k=1}^{T} \left(1-\frac{n}{p}\right)^{T-k} \frac{n}{p}\|\vw_{k}^*-\vw_i^*\|^2
    + \frac{n\sigma^2}{p-n-1}\sum_{k=1}^{T} \left(1-\frac{n}{p}\right)^{T-k} \\
    &-\left(1-\frac{n}{p}\right)^{i} \|\vw_i^*\|^2-\sum_{k=1}^{i} \left(1-\frac{n}{p}\right)^{i-k} \frac{n}{p}\|\vw_{k}^*-\vw_i^*\|^2
    - \frac{n\sigma^2}{p-n-1}\sum_{k=1}^{i} \left(1-\frac{n}{p}\right)^{i-k}
    \Bigg\}\\
    =&\frac{1}{T-1}\sum_{i=1}^{T-1} \Bigg\{\left[\left(1-\frac{n}{p}\right)^T-\left(1-\frac{n}{p}\right)^i\right]\|\vw_i^*\|^2+\sum_{k=1}^{i} \frac{n}{p}\left[\left(1-\frac{n}{p}\right)^{T-k}-\left(1-\frac{n}{p}\right)^{i-k}\right]\|\vw_k^*-\vw_i^*\|^2\\
    &+\sum_{k=i+1}^T \frac{n}{p}\left(1-\frac{n}{p}\right)^{T-k}\|\vw_k^*-\vw_i^*\|^2+\frac{n\sigma^2}{p-n-1} \sum_{k=1}^{i}\left[\left(1-\frac{n}{p}\right)^{T-k}-\left(1-\frac{n}{p}\right)^{i-k}\right]\\
    &+\frac{n\sigma^2}{p-n-1} \sum_{k=i+1}^{T}\left(1-\frac{n}{p}\right)^{T-k}
     \Bigg\}\\
     =&\frac{1}{T-1}\sum_{i=1}^{T-1} \Bigg\{\left[\left(1-\frac{n}{p}\right)^T-\left(1-\frac{n}{p}\right)^i\right]\|\vw_i^*\|^2+\sum_{j>i}^{T} c_{i,j}\|\vw_i^*-\vw_j^*\|^2\\
    &+\frac{n\sigma^2}{p-n-1} \sum_{k=1}^{i}\left[\left(1-\frac{n}{p}\right)^{T-k}-\left(1-\frac{n}{p}\right)^{i-k}\right]+\frac{n\sigma^2}{p-n-1} \sum_{k=i+1}^{T}\left(1-\frac{n}{p}\right)^{T-k}
     \Bigg\}\\
  =&\frac{1}{T-1}\sum_{i=1}^{T-1} \Bigg\{\left[\left(1-\frac{n}{p}\right)^T-\left(1-\frac{n}{p}\right)^i\right]\|\vw_i^*\|^2+\sum_{j>i}^{T} c_{i,j}\|\vw_i^*-\vw_j^*\|^2\\
  &+\frac{n\sigma^2}{p-n-1}\left[\sum_{k=1}^T \left(1-\frac{n}{p}\right)^{T-k}-\sum_{k=1}^i \left(1-\frac{n}{p}\right)^{i-k}\right]\Bigg\}\\
=&\frac{1}{T-1}\sum_{i=1}^{T-1} \Bigg\{\left[\left(1-\frac{n}{p}\right)^T-\left(1-\frac{n}{p}\right)^i\right]\|\vw_i^*\|^2+\sum_{j>i}^{T} c_{i,j}\|\vw_i^*-\vw_j^*\|^2\\
  &+\frac{n\sigma^2}{p-n-1}\left[\frac{1-\left(1-\frac{n}{p}\right)^T}{1-\left(1-\frac{n}{p}\right)}-\frac{1-\left(1-\frac{n}{p}\right)^i}{1-\left(1-\frac{n}{p}\right)}\right]\Bigg\}\\
  =&\frac{1}{T-1}\sum_{i=1}^{T-1} \Bigg\{\left[\left(1-\frac{n}{p}\right)^T-\left(1-\frac{n}{p}\right)^i\right]\|\vw_i^*\|^2+\sum_{j>i}^{T} c_{i,j}\|\vw_i^*-\vw_j^*\|^2\\
  &+\frac{n\sigma^2}{p-n-1} \frac{p}{n}\left[\left(1-\left(1-\frac{n}{p}\right)^T\right)-\left(1-\left(1-\frac{n}{p}\right)^i\right)\right]\Bigg\}\\
 =&\frac{1}{T-1}\sum_{i=1}^{T-1} \Bigg\{\left[\left(1-\frac{n}{p}\right)^T-\left(1-\frac{n}{p}\right)^i\right]\|\vw_i^*\|^2+\sum_{j>i}^{T} c_{i,j}\|\vw_i^*-\vw_j^*\|^2\\
    &+\frac{p\sigma^2}{p-n-1}\left[\left(1-\frac{n}{p}\right)^{i}-\left(1-\frac{n}{p}\right)^{T}\right]
     \Bigg\}\\
     =&\frac{1}{T-1}\sum_{i=1}^{T-1}\Bigg\{(r^T-r^i)\|\vw_i^*\|^2+\sum_{j>i}^T c_{i,j}\|\vw_i^*-\vw_j^*\|^2
     +\frac{p\sigma^2}{p-n-1}\left(r^i-r^T\right)
        \Bigg\}.
\end{align*}

Based on \cref{eq:first}, we can also obtain the exact form of the generalization error. Specifically, \begin{align*}
    &\E[\|\vw_T-\vw_i^*\|^2]\\
    =&\left(1-\frac{n}{p}\right)^T \|\vw_i^*\|^2+\sum_{k=1}^T \frac{n}{p}\left(1-\frac{n}{p}\right)^{T-k}\|\vw_k^*-\vw_i^*\|^2+\frac{n\sigma^2}{p-n-1}\sum_{k=1}^T\left(1-\frac{n}{p}\right)^{T-k},
\end{align*}
such that
\begin{align*}
    \E[G_T]=&\frac{1}{T}\sum_{i=1}^T \E[\|\vw_T-\vw_i^*\|^2]\\
    =&\frac{1}{T}\left(1-\frac{n}{p}\right)^T \sum_{i=1}^T \|\vw_i^*\|^2+\frac{1}{T}\sum_{k=1}^T \frac{n}{p}\left(1-\frac{n}{p}\right)^{T-k} \sum_{i=1}^T \|\vw_k^*-\vw_i^*\|^2 \\
    &+\frac{n\sigma^2}{p-n-1}\sum_{k=1}^T\left(1-\frac{n}{p}\right)^{T-k}\\
    =&\frac{1}{T}\left(1-\frac{n}{p}\right)^T \sum_{i=1}^T \|\vw_i^*\|^2+\frac{1}{T}\sum_{k=1}^T \frac{n}{p}\left(1-\frac{n}{p}\right)^{T-k} \sum_{i=1}^T \|\vw_k^*-\vw_i^*\|^2 \\
    &+\frac{n\sigma^2}{p-n-1}\frac{1-\left(1-\frac{n}{p}\right)^T}{1-\left(1-\frac{n}{p}\right)}\\
    =&\frac{1}{T}\left(1-\frac{n}{p}\right)^T \sum_{i=1}^T \|\vw_i^*\|^2+\frac{1}{T}\sum_{k=1}^T \frac{n}{p}\left(1-\frac{n}{p}\right)^{T-k} \sum_{i=1}^T \|\vw_k^*-\vw_i^*\|^2\\  &+\frac{p\sigma^2}{p-n-1}\left[1-\left(1-\frac{n}{p}\right)^T\right]\\
    =&\frac{r^T}{T}\sum_{i=1}^T \|\vw_i^*\|^2+\frac{1}{T}\sum_{i=1}^T \frac{nr^{T-i}}{p}\sum_{k=1}^T\|\vw_k^*-\vw_i^*\|^2+\frac{p\sigma^2}{p-n-1}\left(1-r^T\right).
\end{align*}




\subsection{Proof of \cref{pro:two_task_back}}

Based on \cref{thm:exactform}, it follows that
\begin{align*}
    \E[F_2]=& (r^2-r)\|\vw_1^*\|^2+\frac{n}{p}\|\vw_1^*-\vw_2^*\|^2+\frac{nr\sigma^2}{p-n-1}\\
    =& -\left(1-\frac{n}{p}\right)\frac{n}{p}\|\vw_{1,s}^*\|^2+\frac{n}{p}\|\vw_{1,s}^*\|^2+\frac{n}{p}\|\vw_{2,s}^*\|^2-2\frac{n}{p}\langle\vw_{1,s}^*,\vw_{2,s}^* \rangle+\frac{nr\sigma^2}{p-n-1}\\
    =& \left(\frac{n}{p}\right)^2 \|\vw_{1,s}^*\|^2+\frac{n}{p}\|\vw_{2,s}^*\|^2-2\frac{n}{p}\langle\vw_{1,s}^*,\vw_{2,s}^* \rangle+\frac{nr\sigma^2}{p-n-1}.
\end{align*}
When $\sigma^2<\frac{p-n-1}{p}\|\vw_1^*\|^2$, 
  \begin{align*}
      \frac{n}{p}\|\vw_1^*\|^2+\|\vw_2^*\|^2+\frac{(p-n)\sigma^2}{p-n-1}\leq \|\vw_1^*\|^2+\|\vw_2^*\|^2,
  \end{align*}
  such that $\E[F_2]\leq 0$ if
    \begin{align*}
      2\langle \vw_{1,\gS_1}^*, \vw_{2,\gS_2}^*\rangle\geq \frac{n}{p}\|\vw_1^*\|^2+\|\vw_2^*\|^2+\frac{(p-n)\sigma^2}{p-n-1}.
  \end{align*}

\subsection{Proof of \cref{pro:order1}}

Without loss of generality, we assume that $\|\vw_i^*-\vw_j^*\|=1$ for task $i$ in Category I and task $j$ in Category II. It follows that
\begin{align*}
    \Tilde{F}_T(\vw_T)=&\sum_{i<i^*}c_{i,i^*}+\sum_{j>i^*}c_{i^*,j}\\
=&(1-r)\left(\sum_{i=1}^{i^*-1}(r^{T-i}-r^{i^*-i}+r^{T-i^*})+\sum_{j=i^*+1}^{T}(r^{T-i^*}-r^{j-i^*}+r^{T-j})\right)\\
=&(1-r)\left((T-1)\cdot r^{T-i^*}+r^{T-i^*+1}\frac{r^{i^*-1}-1}{r-1}-r\frac{r^{i^*-1}-1}{r-1}+1-r^{T-i^*}\right)\\
=& (1-r)(T-2)r^{T-i^*}+(r^{T-i^*}-1)(1-r^{i^*-1})r + (1-r).
\end{align*}

Letting $\alpha \defeq r^{T-i^*}$. Then minimizing $\Tilde{F}_T(\vw_T)$ is equivalent to minimize
\begin{align*}
    &(1-r)(T-2)\alpha+(\alpha -1)(1-\frac{r^{T-1}}{\alpha})r\\
    =&((1-r)(T-2)+r)\alpha+\frac{r^{T}}{\alpha}-r^{T}-r.
\end{align*}

By setting the derivative w.r.t. $\alpha$ to $0$, we can have that the optimal value of $\alpha$ is
\begin{align}
    \alpha = \sqrt{\frac{r^T}{T-2-(T-1)r}}
\end{align}
which is clearly increasing with $r$. Therefore, the optimal order of the special task $i^*$ is non-increasing with $r$, i.e., non-decreasing with $\frac{n}{p}$.

\subsection{Proof of \cref{pro:order2}}

Without loss of generality, we assume that for any task $i$ and $j$
\begin{align*}
     \|\vw_i^*-\vw_j^*\|^2=
     \begin{cases}
        0, & \text{if task $i$ and $j$ are in the same category;}\\
        1, & \text{if task $i$ and $j$ are in the different categories.}
     \end{cases}
\end{align*}

Based on the closed form of forgetting, we can see that it suffices to minimize $\sum_{i=1}^{T-1}\sum_{j>i}^T c_{i,j}\|\vw_i^*-\vw_j^*\|^2$ in order to minimize the forgetting $F_T(\vw_T)$, where $c_{i,j}=(1-r)(r^{T-i}-r^{j-i}+r^{T-j})$. Besides, since whenever we change the order between the $i$-th task and the $j$-th task, the value of $r^{T-i}+r^{T-j}$ does not change. In other words, only the term $r^{j-i}$ affects the optimal task order, which should minimize $\sum_{i=1}^{T-1}\sum_{j>i}^T (-r^{j-i})\|\vw^*_i-\vw_j^*\|^2$.

(1) For the case $T=4$, there are three effective task orders: (1) task $1\in C_1$, task $2\in C_1$, task $3\in C_2$, task $4\in C_2$ ($(C_1,C_1,C_2,C_2)$ for simplicity); (2) $(C_1,C_2,C_1,C_2)$; (3) $(C_1,C_2,C_2,C_1)$. Swapping all tasks in $C_1$ with all tasks in $C_2$ does not change the value of forgetting, e.g., $(C_1,C_1,C_2,C_2)$ has the same forgetting with $(C_2,C_2,C_1,C_1)$. In what follows, we compare $\sum_{i=1}^{T-1}\sum_{j>i}^T (-r^{j-i})\|\vw^*_i-\vw_j^*\|^2$ among these three orders.

(a) For $(C_1,C_1,C_2,C_2)$, 
\begin{align*}
    \sum_{i=1}^{T-1}\sum_{j>i}^T (-r^{j-i})\|\vw^*_i-\vw_j^*\|^2
    =-(r^2+r^3+r+r^2).
\end{align*}
(b) For $(C_1,C_2,C_1,C_2)$,
\begin{align*}
    \sum_{i=1}^{T-1}\sum_{j>i}^T (-r^{j-i})\|\vw^*_i-\vw_j^*\|^2
    =-(r+r^3+r+r).
\end{align*}
(c) For $(C_1,C_2,C_2,C_1)$,
\begin{align*}
    \sum_{i=1}^{T-1}\sum_{j>i}^T (-r^{j-i})\|\vw^*_i-\vw_j^*\|^2
    =-(r+r^2+r+r^2).
\end{align*}
It is clear that the alternating task order, i.e., $(C_1,C_2,C_1,C_2)$ and $(C_2,C_1,C_2,C_1)$, is the optimal order for this special case.

(2) For the case $T=6$, based on the closed form of forgetting in \cref{thm:exactform}, we can use computer programming to show that besides the perfectly alternating task order, i.e., $(C_1,C_2,C_1,C_2,C_1,C_2)$ and $(C_2,C_1,C_2,C_1,C_2,C_1)$, there are 10 effective task orders as illustrated in Table \ref{table.validate_2_groups_6}. We further evaluate the difference of forgetting between each task order in Table \ref{table.validate_2_groups_6} and the perfectly alternating task order, where a positive difference means that the corresponding task order will lead a larger forgetting than the perfectly alternating task order. It can be verified that the difference of forgetting is positive for all the task orders in Table \ref{table.validate_2_groups_6}, which indicates that the optimal task order is the perfectly alternating task order.

\begin{table}[ht!]
\centering
\begin{tabular}{|c | c| c|} 
\hline
\textbf{Index} & \textbf{Order} & \textbf{Difference of forgetting}\\
\hline
1 & $(C_1,C_2,C_1,C_2,C_1,C_2)$ & $0$ \\\hline 
2 & $(C_1,C_1,C_2,C_1,C_2,C_2)$ & $r\left(2-2r+2r^{2}-2r^{3}\right)$ \\\hline   
3 & $(C_1,C_1,C_2,C_2,C_1,C_2)$  & $r\left(2-3r+2r^{2}-r^{3}\right)$ \\\hline    
4 & $(C_1,C_1,C_2,C_2,C_2,C_1)$ & $r\left(3-3r-r^{3}+r^{4}\right)$ \\\hline     
5 & $(C_1,C_2,C_2,C_1,C_1,C_2)$ & $r\left(2-4r+2r^{2}\right)$ \\\hline
6 & $(C_1,C_2,C_2,C_1,C_2,C_1)$ & $r\left(1-2r+2r^{2}-2r^{3}+r^{4}\right)$ \\\hline
7 & $(C_1,C_1,C_1,C_2,C_2,C_2)$ & $r\left(4-2r-2r^{3}\right)$ \\\hline
8 & $(C_1,C_2,C_1,C_2,C_2,C_1)$ & $r\left(1-2r+2r^{2}-2r^{3}+r^{4}\right)$ \\\hline
9 & $(C_1,C_2,C_1,C_1,C_2,C_2)$ & $r\left(2-3r+2r^{2}-r^{3}\right)$ \\\hline    
10 & $(C_1,C_2,C_2,C_2,C_1,C_1)$ & $r\left(3-3r-r^{3}+r^{4}\right)$ \\\hline  
\end{tabular}
\caption{Evaluation of the difference of forgetting between each effective task order and the perfectly alternating task order $(C_1,C_2,C_1,C_2,C_1,C_2)$, where a positive difference means that the corresponding task order will lead a larger forgetting than the perfectly alternating task order.}
\label{table.validate_2_groups_6}
\end{table}

\subsection{Proof of \cref{pro:order3}}

Following the same strategy with Special case II, we can have Table \ref{table.validate_3_groups_6} to show all effective task orders and their difference of forgetting with the  perfectly alternating task order, i.e., $(C_1,C_2,C_3,C_1,C_2,C_3)$ and its `equivalent' task orders (e.g., $(C_1,C_3,C_2,C_1,C_3,C_2)$). It can also be verified that the perfectly alternating task order is the optimal task order in this case.

\begin{table}[ht!]
\centering
\begin{tabular}{|c | c| c|} 
\hline
\textbf{Index} & \textbf{Order} & \textbf{Difference of forgetting}\\
\hline
1 & $(C_1,C_2,C_3,C_1,C_2,C_3)$ & $0$ \\\hline
2 & $(C_1,C_2,C_1,C_2,C_3,C_3)$ & $r\left(1+2r-3r^{2}\right)$ \\\hline
3 & $(C_1,C_2,C_2,C_3,C_3,C_1)$ & $r\left(2-3r^{2}+r^{4}\right)$ \\\hline       
4 & $(C_1,C_2,C_1,C_3,C_2,C_3)$ & $r^{2}\left(2-2r\right)$ \\\hline
5 & $(C_1,C_2,C_3,C_2,C_1,C_3)$ & $r^{2}\left(1-2r+r^{2}\right)$ \\\hline       
6 & $(C_1,C_2,C_3,C_1,C_3,C_2)$ & $r^{2}\left(1-2r+r^{2}\right)$ \\\hline       
7 & $(C_1,C_1,C_2,C_3,C_2,C_3)$ & $r\left(1+2r-3r^{2}\right)$ \\\hline
8 & $(C_1,C_2,C_2,C_1,C_3,C_3)$ & $r\left(2-2r^{2}\right)$ \\\hline
9 & $(C_1,C_1,C_2,C_2,C_3,C_3)$ & $r\left(3-3r^{2}\right)$ \\\hline
10 & $(C_1,C_2,C_1,C_3,C_3,C_2)$ & $r\left(1+r-3r^{2}+r^{3}\right)$ \\\hline     
11 & $(C_1,C_2,C_3,C_3,C_1,C_2)$ & $r\left(1-3r^{2}+2r^{3}\right)$ \\\hline     
12 & $(C_1,C_2,C_3,C_3,C_2,C_1)$ & $r\left(1-2r^{2}+r^{4}\right)$ \\\hline      
13 & $(C_1,C_2,C_2,C_3,C_1,C_3)$ & $r\left(1+r-3r^{2}+r^{3}\right)$ \\\hline    
14 & $(C_1,C_1,C_2,C_3,C_3,C_2)$ & $r\left(2-2r^{2}\right)$ \\\hline
15 & $(C_1,C_2,C_3,C_2,C_3,C_1)$ & $r^{2}\left(2-3r+r^{3}\right)$ \\\hline 
\end{tabular}
\caption{Evaluation of the difference of forgetting between each effective task order and the perfectly alternating task order $(C_1,C_2,C_3,C_1,C_2,C_3)$, where a positive difference means that the corresponding task order will lead a larger forgetting than the perfectly alternating task order.}
\label{table.validate_3_groups_6}
\end{table}

\subsection{Proof of \cref{thm:underparameterized}}\label{proof:underparameterized}

\emph{Intuitive explanation of \cref{thm:underparameterized}:} In the underparameterized region, minimizing the loss  \cref{eq:loss} for the current task $t$ will lead to a unique solution for this task, which does not depend on the learning process and the learned model of previous tasks. That is to say, the task learning is independent among all tasks, such that (i) the learning order of the first $T-1$ tasks does not matter, and (ii)
both forgetting and generalization performance depend only on the model distance between the last task and the other tasks, i.e., $\sum_{i=1}^{T-1}\|\vw_T^*-\vw_i^*\|^2$.

Now we formally prove \cref{thm:underparameterized}.

For the underparameterized regime, the solution of minimizing the training loss is
\begin{align*}
    \vw_t = &(\mX_t\mX_t^{\top})^{-1}\mX_t \vy_t\\
    =& (\mX_t\mX_t^{\top})^{-1}\mX_t \left(\mX_t^{\top} \vw_t^* + \vz_t \right)\\
    =& \vw_t^* + (\mX_t\mX_t^{\top})^{-1}\mX_t \vz_t.
\end{align*}

It follows that
\begin{align*}
    \vw_T - \vw_i^*=\vw_T^*-\vw_i^* + (\mX_T\mX_T^{\top})^{-1}\mX_T \vz_T,
\end{align*}
such that the model error for the $i$-th task can be represented as:
\begin{align*}
    \|\vw_T-\vw_i^*\|^2=\|\vw_T^*-\vw_i^*\|^2+\|(\mX_T\mX_T^{\top})^{-1}\mX_T \vz_T\|^2.
\end{align*}

By taking expectation on both sides,  we can have
\begin{align*}
    \E \|\vw_T-\vw_i^*\|^2 = \|\vw_T^*-\vw_i^*\|^2 + \frac{p\sigma^2}{n - p - 1}.
\end{align*}

Therefore, it can be shown that
\begin{align*}
    \E[G_T]=\E \frac{1}{T}\sum_{i=1}^T \|\vw_T-\vw_i^*\|^2 = \left(\frac{1}{T}\sum_{i=1}^T\|\vw_T^*-\vw_i^*\|^2\right) + \frac{ p\sigma^2}{n - p - 1}
\end{align*}
and
\begin{align*}
    \E[F_T]=&\frac{1}{T-1}\sum_{i=1}^{T-1}\E \left[\|\vw_T-\vw_i^*\|^2-\|\vw_i-\vw_i^*\|^2\right]\\
    =&\frac{1}{T-1}\sum_{i=1}^{T-1}\|\vw_T^*-\vw_i^*\|^2.
\end{align*}

\end{document}